\documentclass[final,3p,times,twocolumn]{elsarticle}
\usepackage{graphicx}
\usepackage{url}
\usepackage{slashbox}
\usepackage{wrapfig}
\usepackage{algorithmic}
\usepackage{algorithm}
\usepackage{float}

\usepackage{amssymb}

\begin{document}

\begin{frontmatter}



\title{A Generalizable Knowledge Framework for Semantic Indoor Mapping Based on Markov Logic Networks and Data Driven MCMC}


\author{Ziyuan Liu\fnref{label2}}
\address{Siemens AG, Corporate Technology, Munich, Germany\\
Institute of Automatic Control Engineering, Technische Universit\"at M\"unchen, Munich, Germany\\ Institute for Advanced Study, Techniche Universit\"at M\"unchen, Lichtenbergstrasse 2a, D-85748 Garching, Germany
 }
\ead{ziyuan.liu@tum.de}
\fntext[label2]{Corresponding author. Postal address: Karlstr. 45, room 5001, 80333, Munich, Germany. Telephone: +49-89-289-26900. Fax: +49-89-289-26913.}

\author{Georg von Wichert}
\ead{georg.wichert@siemens.com}
\address{Siemens AG, Corporate Technology, Munich, Germany \\ Institute for Advanced Study, Techniche Universit\"at M\"unchen, Lichtenbergstrasse 2a, D-85748 Garching, Germany}

\begin{abstract}
In this paper, we propose a generalizable knowledge framework for data abstraction, i.e. finding compact abstract model for input data using predefined abstract terms. Based on these abstract terms, intelligent autonomous systems, such as a robot, should be able to make inference according to specific knowledge base, so that they can better handle the complexity and uncertainty of the real world. We propose to realize this framework by combining Markov logic networks (MLNs) and data driven MCMC sampling, because the former are a powerful tool for modelling uncertain knowledge and the latter provides an efficient way to draw samples from unknown complex distributions. Furthermore, we show in detail how to adapt this framework to a certain task, in particular, semantic robot mapping. Based on MLNs, we formulate task-specific context knowledge as descriptive soft rules. Experiments on real world data and simulated data confirm the usefulness of our framework.

\end{abstract}


\end{frontmatter}


\section{Introduction}
\label{int}
In recent years, the performance of autonomous systems has been greatly improved. Multicore CPUs, bigger RAMs, new sensors, faster data flow and so on have made many applications possible which seemed to be unrealistic in the past. However, the performance of such systems tends to become quite limited, as soon as they leave their carefully engineered operating environments. On the other hand, people may ask, why we humans can handle highly complex problems with relatively small computational power and limited memory (compared with computers). Maybe the exact answer to this question still remains unclear, however, it is obvious that abstraction and knowledge together play an important role. We humans understand the world in abstract terms and have the necessary knowledge, based on which we can make inference given only partially available data. For instance, if a person sees a desk in an office room, instead of memorizing the world coordinates of all the surface points of the desk, he/she will only notice that there is an object ``desk" at a certain position, and even this position is probably described in abstract terms like ``beside the window" or ``near to the door". According to his/her knowledge, this person can make some reasonable assumptions, such as there could be some ``books" in the ``drawer" of the desk, instead of some ``shoes" being inside, without openning the drawer. In our work, we aim to provide autonomous systems the ability to abstract and to infer based on given knowledge so that they can better handle the complexity and uncertainty of the real world.

\indent Knowledge processing has found its successful applications in a very wide range: social network analysis \cite{bingwang2010}, expert system \cite{Fukuda1991}, data mining \cite{Poklemba2011}, business process management \cite{Huang2006}, search engine \cite{Milne2007}, etc. People use knowledge to express their belief on a certain topic, which is learned from their daily life. Such knowledge holds for the most cases, nevertheless, there still exit scenarios where it fails. Thus, it is reasonable to model and use knowledge in the form of soft rules, which allow the existence of contradiction and retain the flexibility by defining knowledge as modular rules. For this purpose, Markov logic networks (MLNs) \cite{richardson2006markov} are a good fit, because they combine first-order logic \cite{barwise1977introduction} and probabilistic graphical models \cite{koller2009probabilistic}. First-order logic is able to compactly present knowledge in formulas (hard rules), and probabilistic graphical models are good at handling uncertainty. The combination of both makes it possible to express knowledge as soft rules (formulas attached with weight indicating uncertainty) in a systematic way. Although MLNs are a quite new method, existing since 2006, many MLNs-based approaches have been proposed so far, such as \cite{jeong2010context}, \cite{leung2011flexible}, \cite{wang2008hybrid}, \cite{nienhuser2011relevance}, etc.

\indent In this paper, we propose a generalizable knowledge framework that provides autonomous systems abstract (semantic) level understanding of the obtained data, using MLNs and Markov Chain Monte Carlo (MCMC) sampling \cite{andrieu2003introduction}. Based on the abstraction, we explicitly make use of knowledge processing to enhance the overall performance of data processing. We define semantic patterns and use them as the fundamental elements of knowledge processing, so that, on this basis, intelligent data processing can be realized. To illustrate the general idea, we focus here on the topic of semantic robot indoor mapping and show the performance of our framework.

\indent The remainder of this paper is structured as follows: in section \ref{srm}, we review related work in the field of semantic robot mapping and list our contributions. In section \ref{gkf}, we explain the fundamental idea of our generalizable knowledge framework. In section \ref{mln}, we briefly introduce the theory of Markov logic networks. In section \ref{oursem}, we detail on how to use our knowledge framework to solve the problem of semantic robot indoor mapping. In section \ref{exp}, we show the performance of our semantic mapping system on real world data and simulated data and compare it with our previous work. In section \ref{sum}, we conclude and give an outlook.

\section{Semantic Robot Mapping}
\label{srm}
\begin{figure*}
	\centering
	\includegraphics[width=1.7\columnwidth]{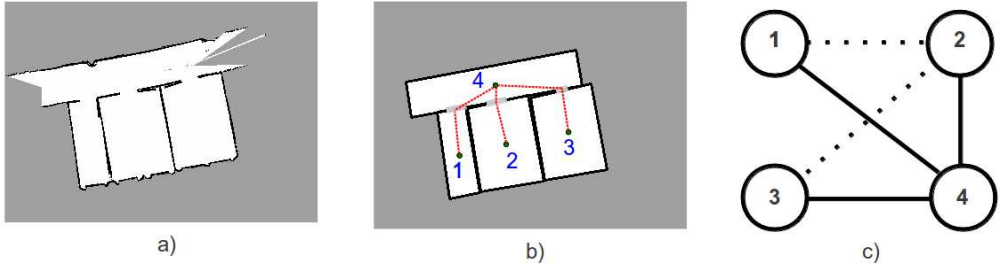}

    \caption{Metric mapping vs. semantic mapping: a) A 2D metric map (occupancy grid map) obtained using \cite{grisetti2007improved}. Such maps are a big matrix of occupancy values and do not provide semantic level information. b) The corresponding semantic map generated by \cite{liu2013extracting}. Basic semantic type ``unit" is introduced, which is represented by a rectangle. Four units are found in the environment. Circles indicate the unit center. Light-gray shows the door. The topology of the map is demonstrated in dashed lines. c) The semantic map represented in pure abstract level. Solid lines indicate that two units are connected by a door, whereas dashed lines connect two neighbour units. }
    \label{figure:example}
\end{figure*}

\indent The goal of robot metric mapping is to build an accurate, globally consistent, and metric map of a robot’s operating environments, so that the robot can localize itself, plan a path and finally navigate towards certain positions. Such mapping systems can be found in the field of Simultaneous Localization and Mapping (SLAM) \cite{grisetti2007improved}. Different from metric mapping, semantic robot mapping aims to construct a semantic map for the environments that the robots work in. The focus of semantic mapping is how to describe the environments on the semantic/abstract level, so as to provide valuable semantic information for higher level applications, such as Human Robot Interaction (HRI) and service robots. An example on the comparison between metric and semantic mapping is illustrated in Fig. \ref{figure:example}.

\subsection{Related Work}
\label{related work}
\indent Although semantic mapping is still a relatively new research field, it has already drawn great interest in the academia. In the state of the art there exist many proposals which can be categorized according to different criteria, such as 2D/3D, indoor/outdoor, single/multiple modality and so on. In the following we review these related work with respect to their output form.

A very big body of literature focuses on semantic place labelling which divides the environment into several regions and attaches each region a semantic label like ``office room" or ``corridor". Park and Song \cite{park2011hybrid} proposed a hybrid semantic mapping system for home environments, using explicitly the door information as a key feature. Combining image segmentation and object recognition, Jebari et. al. \cite{jebari2011multi} extended semantic place labelling with integrated object position. Based on human augmented mapping \cite{topp2006topological}, rooms and hallways are represented as Gaussians to help robot navigate in \cite{Nieto-Granda2010}. Pronobis and Jensfelt \cite{pronobis2012large} integrated multi-modal sensory information, human intervention and some common-sense knowledge to classify places with semantic types. Other examples on semantic place labelling can be found in \cite{goerke2009building}, \cite{krishnan2010visual} and \cite{sjoo2012semantic}.

Different from place labelling,  another big group of work concentrates on labelling different constituents of the perceived environments with semantic tags, such as walls, floors, ceilings of indoor environments, or buildings, roads, vegetations of outdoor environments. In \cite{nuechter2008towards}, a logic-based constraint network describing the relations between different constituents is used for the labelling process in indoor environments. Persson and Duckett \cite{persson2007probabilistic} combined range data and omni-directional images to detect outlines of buildings and nature objects in an outdoor setting. Zhu et. al. \cite{zhu2010segmentation} implemented a semantic labelling system on a vehicle to classify urban scenes, based on range image. \cite{pfingsthorn2011semantic} and \cite{sukhatme2008towards} show the application of semantic constituent labelling in underwater scenarios. Other examples in this category can be found in \cite{shim20113d}, \cite{sakenas2007extraction}, \cite{an2012} and \cite{wolf2008semantic}.

\indent Another category of literature consists of object-based semantic mapping systems which use object as basic representation unit of the perceived environment. Such systems usually adopt point cloud processing (e.g. Point Cloud Library \cite{rusu20113d}) and image processing (e.g. OpenCV \cite{opencv_library}) techniques to model or detect objects, and object features like appearance, shape and 3D locations are often used to represent the objects. Rusu et. al. \cite{rusu2009model} proposed a hybrid semantic object mapping system for household environments (mainly kitchens), based on 3D point cloud data. Objects modelled in this work are those which perform utilitarian functions in the kitchen such as kitchen appliances, cupboards, tables, and drawers. An early example on object-based semantic mapping is shown in \cite{limketkai2005relational}, where a relational object map is proposed for laser-based mobile robot 2D mapping, by modelling geometric primitives like line segments as objects. More examples on object-based semantic mapping can be found in \cite{ranganathan2007semantic}, \cite{correa2009semantic}, \cite{pangercicsemantic} and \cite{masonobject}.

\indent In addition to the three categories mentioned above, there exist also a few systems which adopt explicitly a compact semantic model to represent the perceived environments. In \cite{abdul20113d}, a 3D planar model is proposed for indoor environments based on knowledge of spatial relationship of room surfaces. In our previous work \cite{liu2013extracting} we proposed a generative model for extracting semantic indoor floor plan, based on data driven MCMC \cite{tu2005image}. Similarly, Geiger et. al. \cite{geiger2011generative} introduced a generative model for explaining urban scenes with semantic types, and realized the entire system using MCMC sampling.

\subsection{Our Contributions}
\indent In this paper, we extend our previous work \cite{liu2013extracting} with knowledge processing, and the work as a whole demonstrates a generalizable framework that bridges the gap between abstract reasoning and primitive data processing. The general output of our framework is a compact abstract model of the incoming data, which is constructed based on predefined abstract concepts. Using Markov Logic Networks, we formulate task-specific knowledge base as descriptive rules which increase abstraction performance, ensure modelling flexibility and are able to handle uncertain knowledge at the same time. In addition, we show a systematic way on how to adopt this framework to a certain task, in particular, indoor semantic mapping.

\indent Unlike semantic labelling processes, whose typical output is a map data set with semantic tags, our mapping system outputs a compact semantic model of the perceived environments, which contains rich semantic information and can therefore ease higher-level robotic applications. We realize our system in a generative manner by employing knowledge-enhanced data driven MCMC sampling. Experiments on real world data and simulated data show promising results.


\section{Generalizable Knowledge Framework}
\label{gkf}
\indent According to Bayes' theorem, a main criterion for evaluating how well the extracted abstract model matches with the input data is the posterior probability of the model conditioned on the data $p({Model}|\textrm{Data})$ which can be calculated as follows:
\begin{equation}
p({Model}|\textrm{Data})\propto p(\textrm{Data}|{Model})\cdot p({Model}).
\label{eq:pos}
\end{equation}
Here, the term $p(\textrm{Data}|{Model})$ is usually called likelihood and indicates how probable the observed data set is for different settings of the model. Note that the likelihood is not a probability distribution over the model, and its integral with respect to the model does not (necessarily) equal one \cite{bishop2007machine}. The term $p({Model})$ is the prior describing what kind of models are possible at all. The goal of our abstraction process is then to find the model ${Model}^*$ that best explains the data and meanwhile complies with the prior, which leads to the maximum of the posterior probability:
\begin{equation}
{Model}^*=\arg\!\max_{\!\!\!\!\!\!\!\!\!\!\!{\small{{Model}}}\in\Omega} \, p({Model}|\textrm{Data}),
\label{eq:argmax}
\end{equation}
where $\Omega$ indicates the entire solution space.

\indent In our framework, we propose to find ${Model}^*$ using a knowledge-enhanced data driven MCMC process by encoding task-specific knowledge as descriptive rules. The entire modelling procedure can be divided into the following steps:
\begin{enumerate}
	\item[1)] \textbf{Define abstract model}: Define necessary abstract types and relations (summarized as abstract variables ${Var_{abs}}$) for explaining the input data, based on which the abstract model is constructed. These are also fundamental elements for expressing the knowledge base in MLN. For instance, abstract types could be ``room", ``corridor" and ``hall" for robot indoor mapping, or ``road", ``traffic flow" and ``building" for traffic scene understanding. Abstract relations could be ``adjacent" or ``not related" and so on. Note that the definition of abstract variables depends mainly on the scenario and the needs of the user.
	\item[2)] \textbf{Design data driven MCMC}: Establish the underlying data driven MCMC process which iteratively improves the compact abstract model from a certain initial guess by applying stochastic sampling. This step includes defining MCMC kernels that are needed to change the abstract model and designing scheduling strategies of the kernels. Examples can be found in \cite{liu2013extracting} and \cite{tu2005image}.
	\item[3)] \textbf{Model knowledge}: Define reasonable task-specific knowledge as descriptive rules using MLNs. MLNs take the abstract variables $Var_{abs}$ as input, and their output is used to calculate certain intermediate control variables ${Var_{icv}}$. Then ${Var_{icv}}$ are used to initialize the functions for calculating the prior $p({Model})$. In this sense, the intermediate control variables ${Var_{icv}}$ are certain function of $Var_{abs}$ as shwon in Fig. \ref{figure:icv}.
\begin{figure}[!htb]
	\centering
	\includegraphics[width=.9\columnwidth]{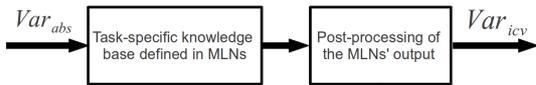}
    \caption{$Var_{icv}=\textrm{Function}(Var_{abs})$}
    \label{figure:icv}
\end{figure}	
	\item[4)] \textbf{Define prior}: Formulate the abstract model as $${Model:=\{Var_{con},Var_{abs},Var_{icv}\},}$$ Then the prior $p({Model})$ is calculated in the following form:
	$$p({Model})=p(Var_{con},Var_{abs},Var_{icv})$$
	\begin{equation}
		=p(Var_{con}|Var_{abs},Var_{icv})\cdot p(Var_{abs},Var_{icv}),
	\label{eq:model}
	\end{equation}
where $Var_{con}$ is the set of continuous variables and should be processed by the functions $p(Var_{con}|Var_{abs},Var_{icv})$ initialized by ${Var_{icv}}$. Since ${Var_{icv}}$ are certain function of ${Var_{abs}}$, we define $p(Var_{con}|Var_{abs},Var_{icv})$ and $p(Var_{abs},Var_{icv})$ as
	\begin{equation}
	p(Var_{con}|Var_{abs},Var_{icv}):=p(Var_{con}|Var_{icv}),
	\end{equation}
	\begin{equation}
	p(Var_{abs},Var_{icv}):=p(Var_{abs}).
	\end{equation}
$p(Var_{abs})$ indicates the probability of different settings of $Var_{abs}$. Note that this term can be designed accordingly, if corresponding knowledge exists, otherwise, it can be considered to follow uniform distribution.
	\item[5)] \textbf{Define likelihood}: Define the likelihood $p(\textrm{Data}|{Model})$ accordingly for calculating the posterior $p({Model}|\textrm{Data})$. A common way to calculate $p(\textrm{Data}|{Model})$ is generating data in the correct formate from the abstract model, and then design $p(\textrm{Data}|{Model})$ on the basis of the comparison between the generated data and the input data.
\end{enumerate}

\indent By establishing this framework, we aim to set up a procedure that systematically makes use of task-specific knowledge in the form of descriptive rules and combines data driven MCMC sampling to extract compact abstract model from input data. Using task-specific knowledge we shape the prior distribution, so that the models that comply with our knowledge have high probability, and other models have low probability. After defining all necessary components, we run the data driven MCMC process to get the correct model.
%
%
\section{Markov Logic Networks}
\label{mln}
\indent Before explaining the theory of Markov Logic Networks (MLNs), we first briefly introduce the two fundamental ingredients of MLNs, which are Markov Networks and First-Order Logic.
\subsection{Markov Networks}
\indent According to \cite{pearl1988probabilistic}, a Markov network is a model for representing the joint distribution of a set of variables $X=(X_1, X_2,\dots,X_n)\in \mathbb{X}$, which constructs an undirected Graph $G$, with each variable represented by a node of the graph. In addition, the model has one potential function $\phi_k$ for each clique in the graph, which is a non-negative real-valued function of the state of that clique. Then the joint distribution represented by a Markov network is calculated as 
\begin{equation}
P(X=x)=\frac{1}{Z}\prod_{k}\phi_k(x_{\{k\}}),
\label{eq:markov}
\end{equation}
with $x_{\{k\}}$ representing the state of the variables in the $k$th clique. The partition function $Z$ is calculated as 
\begin{equation}
Z=\sum_{x\in\mathbb{X}}\prod_k\phi_k(x_{\{k\}}).
\end{equation}
By replacing each clique potential function with an exponentiated weighted sum of features of the state, Markov networks are usually used as log-linear models:
\begin{equation}
P(X=x)=\frac{1}{Z}\exp\left(\sum_{j}\omega_jf_j(X)\right),
\label{eq:markov-2}
\end{equation}
where $f_j(x)$ is the feature of the state and it can be any real-valued function. For each possible state $x_{\{k\}}$ of each clique, a feature is needed with its weight $\omega_j=\log\phi_k(x_{\{k\}})$. Note that for the use of MLNs only binary features are adopted, $f_j(x)\in\{0,1\}.$ For more details on Markov networks, please refer to \cite{pearl1988probabilistic}.

\subsection{First-Order Logic}
Here we briefly introduce some definitions in first-order logic, which are needed to understand the concept of Markov logic networks, for more details on first-order logic, we kindly ask the reader to refer to \cite{genesereth1987logical}.
\begin{itemize}
	\item \textit{Constant} symbols: these symbols represent objects of the interest domain. 
	\item \textit{Variable} symbols: the value of these symbols are the objects represented by the constant symbols. 
	\item \textit{Predicate} symbols: these symbols normally describe relations or attributes of objects. 
	\item \textit{Function} symbols: these symbols map tuples of objects to other objects.
	\item An \textit{atom} or \textit{atomic formula} is a predicate symbol used for a tuple of objects.
	\item A \textit{ground atom} is an atom containing no variables.
	\item A \textit{possible world} assigns a truth value to each possible ground atom. 
	\item Together with logical connectives and quantifiers, a set of logical formulas can be constructed based on atoms to build a \textit{first-order knowledge base}.
\end{itemize}

\subsection{MLNs}
\indent Unlike first-order knowledge bases, which are represented by a set of hard formulas (constraints), Markov logic networks soften the underlying constraints, so that violating a formula only makes a world less probable, but not impossible (the fewer formulas a world violates, the more probable it is). In MLNs, each formula is assigned a weight representing how strong this formula is. According to \cite{richardson2006markov}, the definition of a MLN is:

\textit{
A Markov logic network $L$ is a set of pairs ($F_i,\omega_i$), where $F_i$ is a formula in first-order logic and $\omega_i$ is a real number. Together with a finite set of constants $C=\{c_1,c_2,\dots,c_{|C|}\}$, it defines a Markov network $M_{L,C}$ (equations (\ref{eq:markov}) and (\ref{eq:markov-2})) as follows:
\begin{enumerate}
	\item[1.] $M_{L,C}$ contains one binary node for each possible grounding of each predicate appearing in $L$. The value of the node is 1 if the ground atom is true, and 0 otherwise.
	\item[2.] $M_{L,C}$ contains one feature for each possible grounding of each formula $F_i$ in $L$. The value of this feature is 1 if the ground formula is true, and 0 otherwise. The weight of the feature is the $\omega_i$ associated with $F_i$ in $L$.
\end{enumerate}
}
 
\indent The probability over possible worlds $x$ specified by the ground Markov network $M_{L,C}$ is calculated as 
\begin{eqnarray}
P(X=x)=\frac{1}{Z}\exp\left(\sum_i\omega_i n_i(x)\right)\nonumber\\
=\frac{1}{Z}\prod_i\phi_i(x_{\{i\}})^{n_i(x)},
\end{eqnarray} 
where $n_i(x)$ is the number of true groundings of $F_i$ in $x$, $x_{\{i\}}$ is the state (truth values) of the atoms appearing in $F_i$, and $\phi_i(x_{\{i\}})=e^{\omega_i}$. For more details on MLN, please refer to \cite{richardson2006markov}.

\section{Our Semantic Indoor Mapping System}
\label{oursem}
\begin{figure*}
	\centering
	\includegraphics[width=1.7\columnwidth]{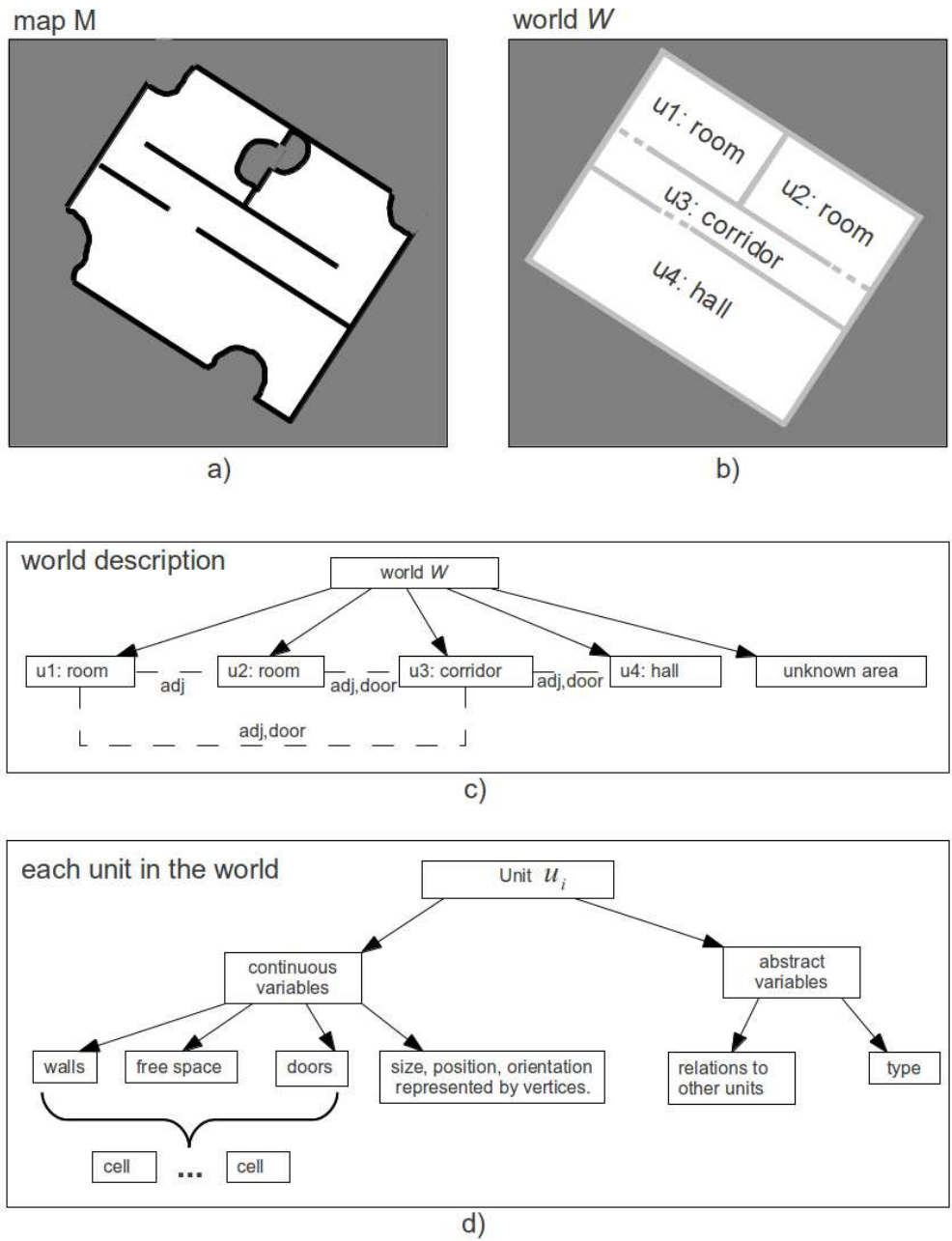}
    \caption{System overview of our semantic mapping system. a) Input occupancy grid map (gray=unknown, white=free, black=occupied). b) The corresponding semantic world model (gray=unknown, white=free, solid light-gray=walls, dashed light-gray=doors) obtained using our system, while assuming the fundamental units have rectangle shape (represented by their four vertices). The units are explained as room, corridor and hall. c) The semantic world described as a scene graph. Two types of relation between units are defined: ``adjacent" and ``irrelevant". If two units share a wall, then they are adjacent (e.g. $u_1$ and $u_3$); otherwise they are irrelevant (e.g. $u_2$ and $u_4$). In addition, connectivity between two units through a door is also detected. d) Each unit in the world contains continuous and abstract variables. The four edges of each unit are its walls. Doors are small line segments comprised of free cells that are located on walls and connect to another unit. All the cells within a unit are considered to belong to free space of the unit. Size, position and orientation of each unit are implicitly represented by its four vertices.}
    \label{figure:system overview}
\end{figure*}

\indent In this section we show in detail how to instantiate and adopt our framework for the task ``semantic robot indoor mapping". Our semantic mapping system aims to build a compact abstract model of indoor environments while taking occupancy grid maps as input data. Such maps can be easily obtained using SLAM process like \cite{grisetti2007improved}. An overview of our system is illustrated in Fig. \ref{figure:system overview}. 

\subsection{An abstract model for indoor environments}
\indent Our abstract model should explain indoor environments in terms of basic indoor space types: ``room", ``corridor" and ``hall", and we denote it as $W$:
\begin{equation}
W:=\{U,T,R\},
\end{equation}
where $U=\{u_i|i=1,\dots, n\}$ represents the set of all $n$ units. Each unit $u_i$ has a rectangle shape and is represented by its four vertices $V_i=\{v_{u_i:j},j\in\{1,2,3,4\}\}$.     $T=\{t_i|i=1,\dots, n\}$ is the set of type of each individual unit, with $t_i\in\{\textrm{room,corridor,hall}\}$. $R=\{r_{p,q}|p=1,\dots, n;q=1,\dots, n\}$ is a $n\times n$ matrix, whose element $r_{p,q}$ describes the relation between the unit $u_p$ and the unit $u_q$, with $r_{p,q}=r_{q,p}\in\{\textrm{adjacent,irrelevant}\}$. If two units share a wall, we define their relation as ``adjacent", otherwise ``irrelevant". By default, we define a unit $u_p$ is irrelevant to itself, i.e. $r_{p,p}=\textrm{irrelevant}$. An example of four units and their relations are depicted in Fig. \ref{figure:system overview}-b and \ref{figure:system overview}-c, where $R$ has the following value:
%
\begin{equation}
    R=
    \left[\begin{array}{cccc}
    \textrm{irr}&\textrm{adj}&\textrm{adj}&\textrm{irr}\\
    \textrm{adj}&\textrm{irr}&\textrm{adj}&\textrm{irr}\\
    \textrm{adj}&\textrm{adj}&\textrm{irr}&\textrm{adj}\\
    \textrm{irr}&\textrm{irr}&\textrm{adj}&\textrm{irr}\\
    \end{array}\right].
\end{equation}
In the following, we call each instance of the abstract model a ``semantic world" or ``world".
\subsection{Likelihood definition}
\indent Let $c(x,y)$ be the grid cell with the coordinate $(x,y)$ in the input occupancy map $M$, then we define the likelihood $p(M|W)$ as follows:
\begin{equation}
p(M|W)=\prod\limits_{c(x,y)\in M}  \alpha(c(x,y))\cdot \beta(c(x,y)).
\label{eq:likelihood}
\end{equation}
Here $\alpha(c(x,y))$ penalizes overlap between units and is given by
\begin{equation}
\alpha(c(x,y))=\psi^{\gamma(c(x,y))},
\end{equation}
with
\begin{eqnarray}
\gamma(c(x,y))&=&\left\{\begin{array}{lc}
\sigma(c(x,y))-1,\sigma(c(x,y))>1\\
0,\textrm{otherwise},\\
\end{array}
\right. 
\end{eqnarray}
where $\psi$ is a penalization factor with $\psi\in(0,1)$. $\sigma(c(x,y))$ indicates the number of units, to which $c(x,y)$ belongs. If there is no overlap in one cell $c(x,y)$, then $\sigma(c(x,y))$ is equal to 0 or 1, in which case $\gamma(c(x,y))$ is 0 (no penalization in cell $c(x,y)$). Otherwise, if $\sigma(c(x,y))$ is bigger than 1, which means the cell $c(x,y)$ belongs to more than one unit, then $\gamma(c(x,y))$ is bigger than 0 (penalization in cell $c(x,y)$).

\indent In equation (\ref{eq:likelihood}), the term $\beta(c(x,y))$ evaluates the match between the world model $W$ and input map $M$, and it is defined as
\begin{equation}
\beta(c(x,y)) = p(c(x,y)|W).
\label{eq:prod}
\end{equation}
For calculating $p(c(x,y)|W)$, we first discretize the cell state $M(x,y)$ of the input map by classifying the occupancy values into three classes 
``occupied=2", ``unknown=1" and ``free=0" so as to generate the classified map $C_M(x,y)$ according to:
\begin{equation}\label{equ:classify}
C_M(x,y)=\left\{\begin{array}{lcc}
2,\quad 0\leq M(x,y)\leq h_o,\\
1, \quad h_o<M(x,y)\leq h_u,\\
0,\quad h_u<M(x,y),
\end{array}
\right.
\end{equation}
where $h_o$ and $h_u$ are the intensity thresholds for occupied and unknown grid cells. Based on our world model $W$ we can also predict expected cell states $C_W(x,y)$ accordingly:
\begin{equation}\label{equ:classify2}
C_W(x,y)=\left\{\begin{array}{lcc}
2,\quad (x,y)\in S_w,\\
1,\quad (x,y)\in S_u,\\
0,\quad (x,y)\in S_f,
\end{array}
\right.
\end{equation}
where $S_w, S_u$ and $S_f$ are the set of all wall cells, unknown cells and free space cells in the world $W$ respectively. $p(c(x,y)|W)$ can then be represented in the form of a lookup-table.

\begin{table}[htb]
	\centering
	\begin{tabular}{|c||*{3}{c|}}\hline
	\backslashbox{$C_W(x,y)$}{$C_M(x,y)$}
	&\makebox[2em]{0}&\makebox[2em]{1}&\makebox[2em]{2}\\\hline\hline
	0 &0.8&0.1&0.1\\\hline
	1 &0.1&0.8&0.1\\\hline
	2&0.1&0.1&0.8\\\hline
	\end{tabular}
	\caption{An example of the look-up table $p(c(x,y)|W)$.}
	\label{TAB:mapping table}
\end{table}

In principle the term $p(c(x,y)|W)$ plays the role of a sensor model. In our case it captures the quality of the original mapping algorithm producing the grid map (including the sensor models for the sensors used during the SLAM process). An example of the look-up table is given in Table \ref{TAB:mapping table}.
\subsection{Prior definition and knowledge processing}
\indent As mentioned in section \ref{gkf}, we need some intermediate control variables $Var_{icv}$, which are output of knowledge processing based on MLNs, to incorporate task-specific knowledge for calculating the prior $p(W)$. So far we defined the abstract model as $W=\{U,T,R\}$, and now we extend it to $$W=\{U,T,R,\Theta\},$$ with $\Theta$ representing the set of intermediate control variables. Then the prior $p(W)$ is given by
\begin{eqnarray}
p(W)&=&p(U,T,R,\Theta)\nonumber\\
&=&p(U|T,R,\Theta) \cdot p(T,R,\Theta)
\end{eqnarray}
where the term $p(U|T,R,\Theta)$ and $p(T,R,\Theta)$ correspond to $p(Var_{con}|Var_{abs}$, $Var_{icv})$ and $p(Var_{icv}, Var_{abs})$ respectively, as described in section \ref{gkf}. $p(U|T,R,\Theta)$ are the functions for processing the continuous variables of the underlying units (see Fig. \ref{figure:system overview}-d), which should be initialized by $\Theta$. In our case, the continuous variables of a unit include size, position, orientation and other constituents (walls, free space and so on), which are implicitly represented by the four vertices of this unit.

\begin{figure}
	\centering
	\includegraphics[width=\columnwidth]{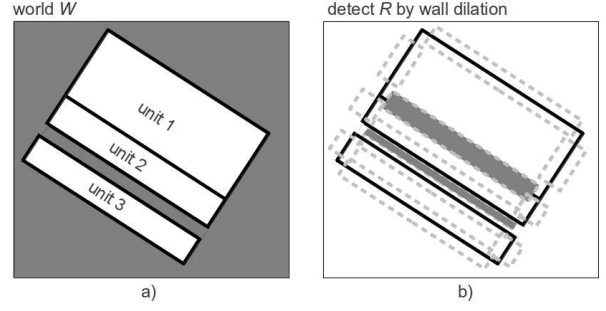}
    \caption{An example on $R$ detection. a) A semantic world $W$ containing three units (black=wall, white=free, gray=unknown). b) All four walls of each unit are dilated, with dashed rectangles in light-gray representing the dilated walls. Overlap of the dilated walls are shown in dark-gray which indicate the relation of ``adjacent". Overlap is detected using connected-components analysis \cite{chang04}. In this example, unit 1 and unit 3 are irrelevant; unit 2 and unit 3 are adjacent; unit 1 and unit 2 are adjacent.}
    \label{figure:detect-r}
\end{figure}

\indent Before we can start processing task-specific knowledge in MLN, two prerequisites need be fulfilled, which are assigning each unit $u_i$ a type $t_i$ and detecting the relations $R$. In order to finish the first, we adopt a hand crafted classifier that categorizes a unit into $room, corridor$ or $hall$ according to its size and length/width ratio. The general idea of this classifier is shown in Table \ref{TAB:classifier}.
\begin{table}[htb]
	\centering
	\begin{tabular}{|c||*{3}{c|}}\hline
	\backslashbox{size}{ratio}
	&\makebox[6em]{small}&\makebox[6em]{big}\\\hline\hline
	small&room&corridor\\\hline
	big&hall&hall\\\hline
	
	\end{tabular}
	\caption{The general idea on how to classify the units.}
	\label{TAB:classifier}
\end{table}
$R$ detection is done based on primitive computational vision techniques: we first dilate all four walls of each unit, and then relation $r_{p,q}$ for the unit $u_p$ and $u_q$ is decided according to connected-components analysis \cite{chang04}. An example of $R$ detection is depicted in Fig.\ref{figure:detect-r}, where $R$ has the following value
\begin{equation}
    R=
    \left[\begin{array}{ccc}
    \textrm{irr}&\textrm{adj}&\textrm{irr}\\
    \textrm{adj}&\textrm{irr}&\textrm{adj}\\
    \textrm{irr}&\textrm{adj}&\textrm{irr}\\
    \end{array}\right].
\end{equation}

\indent Knowledge processing in MLN takes the abstract variables $T,R$ as input in form of evidence predicates, and output of MLN is used to calculate $\Theta$. Necessary predicates (evidence and query) are defined in Table \ref{TAB:evidence} and Table \ref{TAB:query}. Formulas representing knowledge are defined in Table \ref{TAB:formulas}. Given the evidence and the defined formulas, the query probability $p(\textit{SaLe}(u_p,u_q)|R,T)$ is outputted by MLN and is used to calculate $\Theta$. At the current stage, we define $\Theta$ as a matrix similar to $R$, whose element $\theta_{p,q}$ takes a binary value ($true$ or $false$) describing whether two units $u_p,u_q$ should have a wall with the same length. $\theta_{p,q}$ is given by
\begin{equation}
\theta_{p,q}=\left\{\begin{array}{lcc}
true,~~ p(\textit{SaLe}(u_p,u_q)|R,T)>\textrm{threshold}\\\quad ~~~~~~\textrm{and}~~p\neq q,\\
false,~~ \textrm{otherwise}.
\end{array}
\right.
\end{equation}
$\Theta$ for the semantic world shown in Fig. \ref{figure:system overview}-b should be 
\begin{equation}
    \Theta=
    \left[\begin{array}{cccc}
    \textrm{false}&\textrm{true}&\textrm{false}&\textrm{false}\\
    \textrm{true}&\textrm{false}&\textrm{false}&\textrm{false}\\
    \textrm{false}&\textrm{false}&\textrm{false}&\textrm{false}\\
        \textrm{false}&\textrm{false}&\textrm{false}&\textrm{false}
    \end{array}\right].
\end{equation}

\indent Since $\Theta$ is a function of $R$ and $T$, we define $p(U|T,R,\Theta)$ and $p(T,R,\Theta)$ as follows:
\begin{eqnarray}
p(U|T,R,\Theta):=p(U|\Theta),\\
p(T,R,\Theta):=p(T,R).
\end{eqnarray}
Then $p(U|\Theta)$ can be given by
\begin{equation}
p(U|\Theta):=\prod_{p,q\in n}b(u_p,u_q),
\end{equation}
with 
\begin{equation}
b(u_p,u_q)=\left\{\begin{array}{lcc}
e^{-\frac{d}{2\sigma^2}},~~ \theta_{p,q}=true,\\
1,~~ \theta_{p,q}=false, 
\end{array}
\right.
\end{equation}
where $n$ is the total number of units, and $d$ represents the length difference of the neighbour walls of two adjacent units. $e^{-\frac{d}{2\sigma^2}}$ indicates a Gaussian function with mean at zero. Currently we consider $p(T,R)$ follows a uniform distribution. 

\begin{figure}[!htb]
	\centering
	\includegraphics[width=.8\columnwidth]{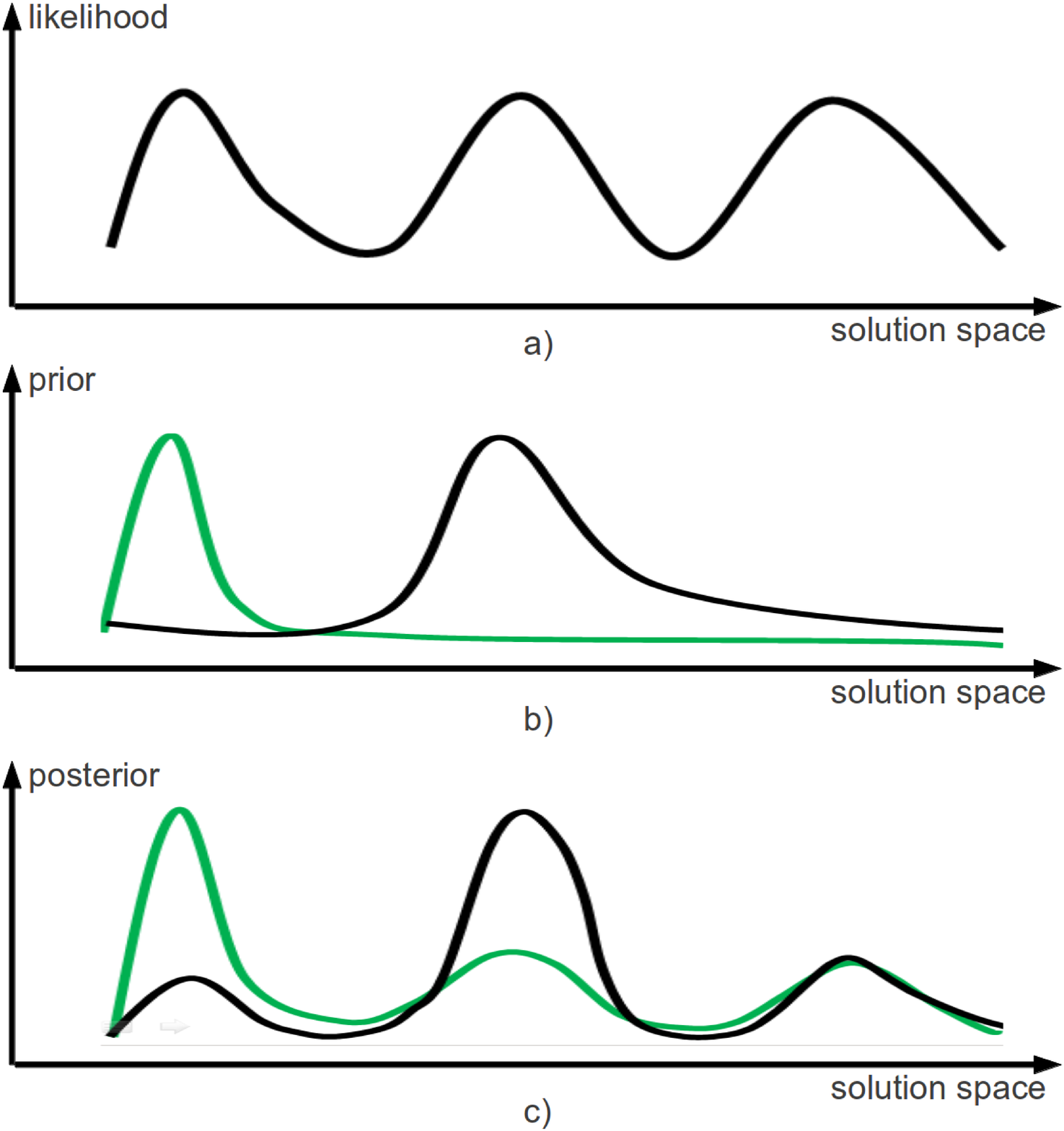}
    \caption{The general concept of knowledge processing illustrated using a one-dimensional example. a) The likelihood for different settings of model, which contains three optima. b) The prior distribution represented by the knowledge base realized in MLN. Different knowledge bases (set of rules) represent different prior distributions (green and black). If no knowledge base is incorporated, it is the same as implementing a knowledge base that represents a uniform distribution which does not influence the posterior, i.e. posterior is only proportional to likelihood. c) Corresponding posterior distributions obtained using the two prior distributions shown in figure b. By setting prior distribution through knowledge processing, we shape the posterior so that the number of optima decreases, which means, the models complying with our knowledge have high prior probability and tend to have high posterior.}
    \label{figure:posterior concept}
\end{figure}

\indent So far, the prior $p(W)$ is defined based on the result of knowledge processing in MLN, which enforces that only the models that comply with the knowledge base have high prior. The general idea of this concept is illustrated in Fig. \ref{figure:posterior concept}.

\begin{table}[htb]
\centering
\begin{tabular}{|ll|}
\hline predicate&explanation \\ \hline
\textit{Room($u_p$)}&Unit $u_p$ has the type of room.\\
\textit{Corr($u_p$)}&Unit $u_p$ has the type of corridor.\\
\textit{Hall($u_p$)}&Unit $u_p$ has the type of hall.\\
\textit{Adj($u_p$,$u_q$)}&Unit $u_p$ and $u_q$ are adjacent.\\
\textit{Irr($u_p$,$u_q$)}&Unit $u_p$ and $u_q$ are irrelevant.\\
\hline 
\end{tabular} 
\caption{Definition of evidence predicates. Abstract variables $T,R$ are represented by these predicates and used as input of knowledge processing in MLN.}
\label{TAB:evidence}     
\end{table}

\begin{table}[htb]
\centering
\begin{tabular}{|ll|}
\hline predicate&explanation \\ \hline
\textit{SaLe($u_p$,$u_q$)}&Unit $u_p$ and $u_q$ have each a \\&wall with the same length.\\
\hline 
\end{tabular} 
\caption{Definition of query predicates. Given the evidence and the defined formulas, MLN can output the query probability $p(\textit{SaLe}(u_p,u_q)|R,T)$.}
\label{TAB:query}     
\end{table}

\begin{table*}[htb]
\centering
\begin{tabular}{|lll|}
\hline index&weight&formula\\ \hline
1&$\infty$&$Irr(u_p,u_q) \to Irr(u_q,u_p)$\\
2&$\infty$&$Adj(u_p,u_q) \to Adj(u_q,u_p)$\\
3&$\infty$&$\textit{SaLe}(u_p,u_q) \to \textit{SaLe}(u_q,u_p)$\\
4&$\infty$&$Irr(u_p,u_q) \to \neg Adj(u_p,u_q)$\\
5&$\omega_5$&$Room(u_p)\land Room(u_q) \land Adj(u_p,u_q)\to \textit{SaLe}(u_p,u_q) $\\
6&$\omega_6$&$Room(u_p)\land Hall(u_q) \land Adj(u_p,u_q)\to \neg \textit{SaLe}(u_p,u_q) $\\
7&$\omega_7$&$Room(u_p)\land Corr(u_q) \land Adj(u_p,u_q)\to \neg \textit{SaLe}(u_p,u_q) $\\
8&$\omega_8$&$Irr(u_q,u_p)\to \neg \textit{SaLe}(u_p,u_q)$\\

\hline 
\end{tabular} 
\caption{Task-specific knowledge defined in MLN. Formulas 1-4 have infinity as weight, which means that these are hard formulas describing certain unbreakable rules, such as symmetry and exclusivity. Formula 5 simply describes the reasonable knowledge that the neighbour walls of two adjacent rooms have same length. Formulas 6-8 just describe the conditions, in which two units should not share a wall with the same length. The weights $\omega_{5,6,7,8}$ can either be learned or manually designed, and examples can be found in \cite{lowd2007efficient} and \cite{jainknowledge}.}
\label{TAB:formulas}     
\end{table*}

\subsection{Design of data driven MCMC}
\indent Having defined the likelihood $p(M|W)$ and the prior $p(W)$, the posterior $p(W|M)$ is given by 
\begin{equation}
p(W|M) \propto p(M|W)\cdot p(W).
\end{equation}
Then our goal is to find the best world $W^*$ that leads to the maximum of posterior probability:
\begin{equation}
W^*=\arg\!\max_{\!\!\!\!\!\!\!\!\!\!\!_{W\in\Omega}} \, p(W|M),
\label{eq:argmax-w}
\end{equation}
with $\Omega$ being the solution space.

\indent For solving equation~(\ref{eq:argmax-w}) we need to efficiently search the large and complexly structured solution space $\Omega$. Here we adopt the approach of \cite{tu2005image}, in which a data driven MCMC technique is proposed for this purpose. The basic idea is to construct a Markov chain that generates samples $W_i$ from the solution space $\Omega$ according to the distribution $p(W|M)$ after some initial burn-in time. One popular approach to construct such a Markov chain is the Metropolis-Hastings (MH) algorithm \cite{chib1995understanding}. In MCMC techniques the Markov chain is constructed by sequentially executing state transitions (in our case from a given world state $W$ to another state $W'$) according to a transition distribution $\Phi(W'|W)$ of the kernels. In order for the chain to converge to a given distribution, it has to be reversible and ergodic~\cite{bishop2007machine}. The MH algorithm achieves this by generating new samples in three steps. First a transition is proposed according to $\Phi(W'|W)$, subsequently a new sample $W'$ is generated by a proposal distribution $Q(W'|W)$, and then it is accepted with the following probability:

\begin{equation}
\lambda(W,W') = \min\left( 1, \frac{p(W'|M) Q(W|W')}{p(W|M) Q(W'|W)} \right)
\label{eq:MH}
\end{equation}

The resulting Markov chain can be shown to converge to $p(W|M)$. However the selection of the proposal distribution is crucial for the convergence rate. Here, we follow the approach of \cite{tu2005image} to propose state transitions for the Markov chain using discriminative methods for the bottom-up detection of relevant environmental features (e.g. walls, doors and so on) and constructing the proposals based on these detection results. More details on realization of the data driven MCMC process, treatment of doors, transition distribution $\Phi(W'|W)$ and proposal distribution $Q(W'|W)$ can be found in our previous work \cite{liu2013extracting}.

\indent In order to design the Markov chain in form of the Metropolis-Hastings algorithm, the kernels that modify the structure of the world are arranged to be reversible. In addition to the four reversible kernel pairs that are defined in our previous work \cite{liu2013extracting}, we propose here a new reversible kernel ``INTERCHANGE" that changes two adjacent units at the same time. The kernels that are currently in use include:
\begin{itemize}
	\item ADD or REMOVE one unit.
		\begin{itemize}
			\item ADD: draw one new unit from certain candidates, then try to add this unit to the world.
			\item REMOVE: try to cancel one existing unit from the world.
		\end{itemize}
	\item SPLIT one unit or MERGE two units.
		\begin{itemize}
			\item SPLIT: try to decompose one existing unit into two units.
			\item MERGE: try to combine two existing units, and generate one new unit out of them.
		\end{itemize}
	\item SHRINK or DILATE one unit.
		\begin{itemize}
			\item SHRINK: try to move one wall of one unit along certain orientation, so that the unit becomes smaller.
			\item DILATE: similarly to SHRINK, move one wall of one unit, so that the unit becomes bigger.
		\end{itemize}
	\item ALLOCATE or DELETE one door
		\begin{itemize}
			\item ALLOCATE: try to attach a door to two existing units.
			\item DELETE: cancel one assigned door.
		\end{itemize}
	\item INTERCHANGE two units: try to change the structure of two adjacent units at the same time, without changing the total size of the two units.
\end{itemize}

\begin{figure*}[!htb]
	\centering
  	\includegraphics[height=1.2\columnwidth]{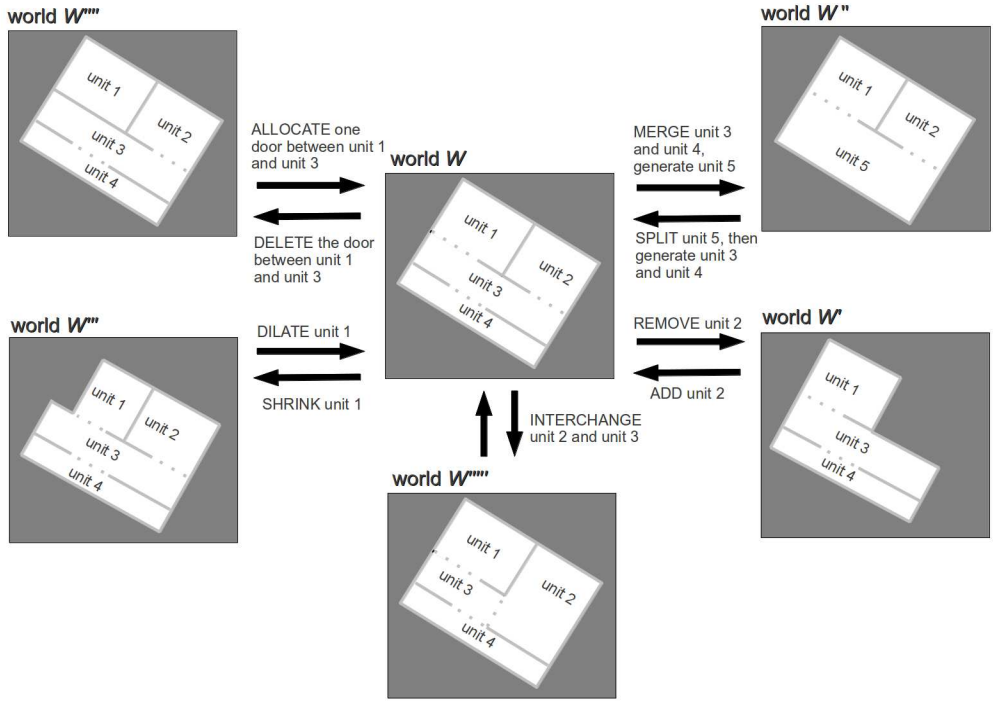}.
    \caption{Reversible MCMC kernels: ADD/REMOVE, SPLIT/MERGE, SHRINK/DILATE, ALLOCATE/DELETE and INTERCHANGE.}
   	\label{figure:mcmc-kernels-knowledge}
\end{figure*}

Fig. \ref{figure:mcmc-kernels-knowledge} shows an example of the reversible MCMC kernels. The world $W$ can transit to $W^{'}$, $W^{''}$, $W^{'''}$, $W^{''''}$ and $W^{'''''}$ by applying the kernel REMOVE, MERGE, SHRINK, DELETE and INTERCHANGE, respectively. By contrast, the world $W^{'}$, $W^{''}$, $W^{'''}$, $W^{''''}$ and $W^{'''''}$ can also transit back to $W$ using corresponding reverse kernel.

\section{Experiments and Discussions}
\label{exp}
\indent In this paper, we extend our previous work with knowledge processing and better bottom-up feature detectors. The performance of our current system is evaluated on various data sets, which include publicly available benchmark data obtained from the Robotics Data Set Repository (Radish)~\cite{Radish}, data acquired using our own mobile robot and simulated data of open source simulators.

\subsection{Evaluation using publicly available benchmark data}
Without loss of generality, we evaluate our current system using publicly available benchmark data. These data are real world data and were acquired with real robots by various researchers. In the following we show the performance of our system on two data sets obtained from \cite{Radish}.

\begin{figure*}[!htb]
	\centering
	\includegraphics[width=1.8\columnwidth]{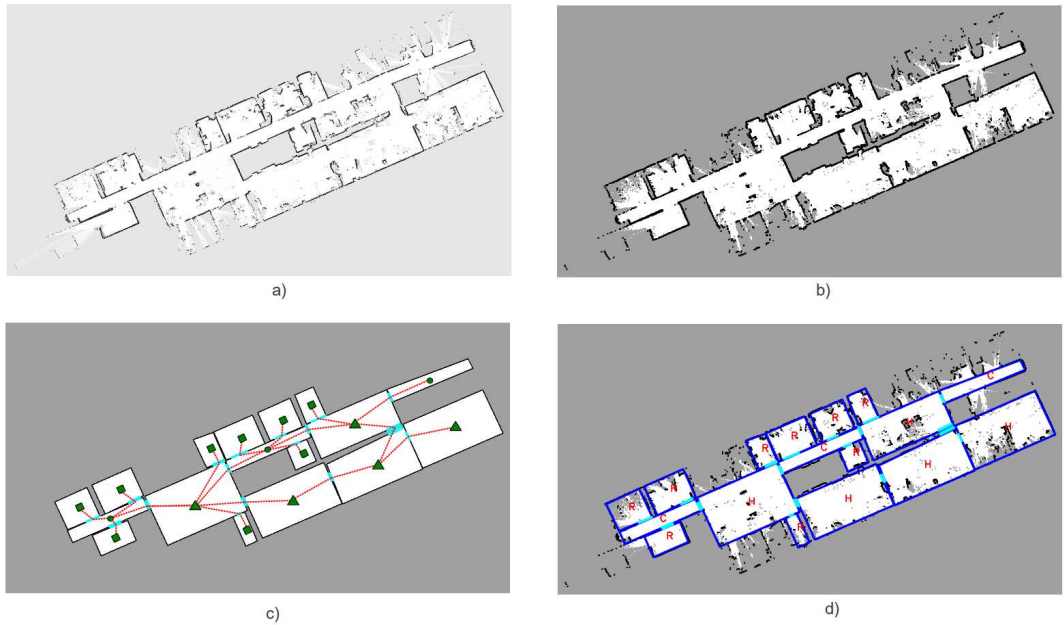}
    \caption{The overall performance of our semantic mapping system. a) Original occupancy grid map $M$ obtained from \cite{Radish}. b) The corresponding classified map $C_M$ (black=occupied, gray=unknown, white=free) obtained according to equation (\ref{equ:classify}). c) Our semantic world $W^*$ (black=wall, gray=unknown, white=free). Connectivity through doors is shown by dashed lines, with cyan representing detected doors. Small triangles, circles and rectangles show the geometric center of hall, corridor and room. d) Here, we plot the world $W^*$ directly onto the classified map $C_M$ to give an intuitive comparison (black=occupied, blue=wall, gray=unknown, white=free, cyan=door). The type of each unit is shown at its center (R=room, H=hall, C=corridor).}
    \label{figure:overall performance}
\end{figure*}

\begin{figure*}[!htb]
	\centering
	\includegraphics[width=1.8\columnwidth]{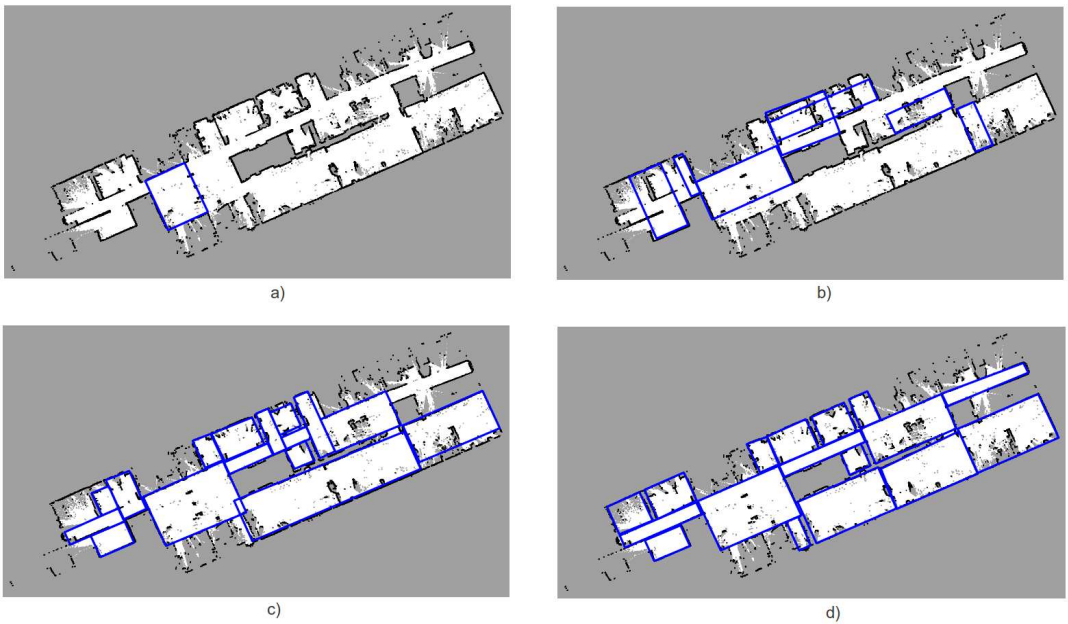}
    \caption{The process of data driven MCMC starts from a random initial guess (figure a). By applying the kernels shown in Fig. \ref{figure:mcmc-kernels-knowledge}, the semantic world model is adapted to the input map, some intermediate results are demonstrated in figure b, c and d. }
    \label{figure:mcmc-steps}
\end{figure*}

\begin{figure*}[!htb]
	\centering
	\includegraphics[width=1.8\columnwidth]{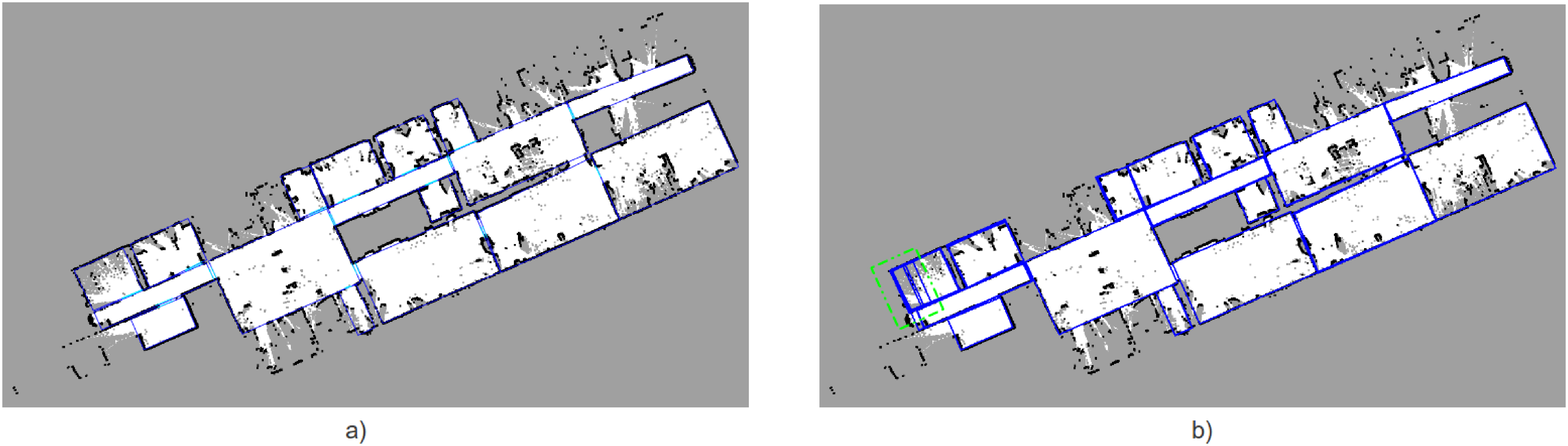}
    \caption{a) After the underlying Markov chain has converged, the world $W^*$ is obtained. Here we purposefully plot the world using very thin blue lines so that the corresponding distribution built by multiple worlds can be better seen. b) The posterior distribution $p(W|M)$ illustrated by 1000 semantic worlds obtained after getting $W^*$ (figure a). Except small variations highlighted by the green dashed rectangle, these 1000 semantic worlds are almost the same, which indicates, that the Markov chain stays stable and that the convergence is well retained.}
    \label{figure:distribution1}
\end{figure*}

Fig. \ref{figure:overall performance} shows the performance of our current system on a big data set. As input, the occupancy grid map $M$ (Fig. \ref{figure:overall performance}-a) of an entire floor of a building is used. Each cell of $M$ is illustrated by its occupancy value that indicates only how probable this cell is occupied. In this sense, this map itself is a big matrix (``1237$\times$672") containing certain continuous values. By applying equation (\ref{equ:classify}), $M$ is thresholded to generate the classified map $C_M$ (Fig. \ref{figure:overall performance}-b), whose cell is described by an abstract state ranging over $\{occupied,unknown,free\}$. Starting from a random initial guess, the semantic world $W$ is adapted to better match the input map $M$ by stochastically applying the kernels shown in Fig. \ref{figure:mcmc-kernels-knowledge}. An example on the process of data driven MCMC is depicted in Fig. \ref{figure:mcmc-steps}. Finally, we get the correct semantic world $W^*$ as shown in Fig. \ref{figure:overall performance}-c. This semantic world $W^*$  comprised of 17 units, each of which is represented by a rectangle, is a compact abstract model of the input map. Not only does $W^*$ accurately represent the geometry of the input map (see Fig. \ref{figure:overall performance}-d), but also $W^*$ provides valuable abstract information for high-level reasoning, such as unit type ($room, corridor, hall$) and connectivity through doors. Fig. \ref{figure:distribution1} shows the posterior distribution $p(W|M)$ built by 1000 samples after the underlying Markov chain has converged, i.e. $W^*$ obtained. In Fig. \ref{figure:distribution1}-b, we can see that except some small variations (highlighted by the green dashed rectangle) these 1000 semantic worlds are almost the same, which indicates, that the whole Markov chain stays stable and that the convergence is well retained.

\begin{figure*}[!htb]
	\centering
	\includegraphics[width=1.8\columnwidth]{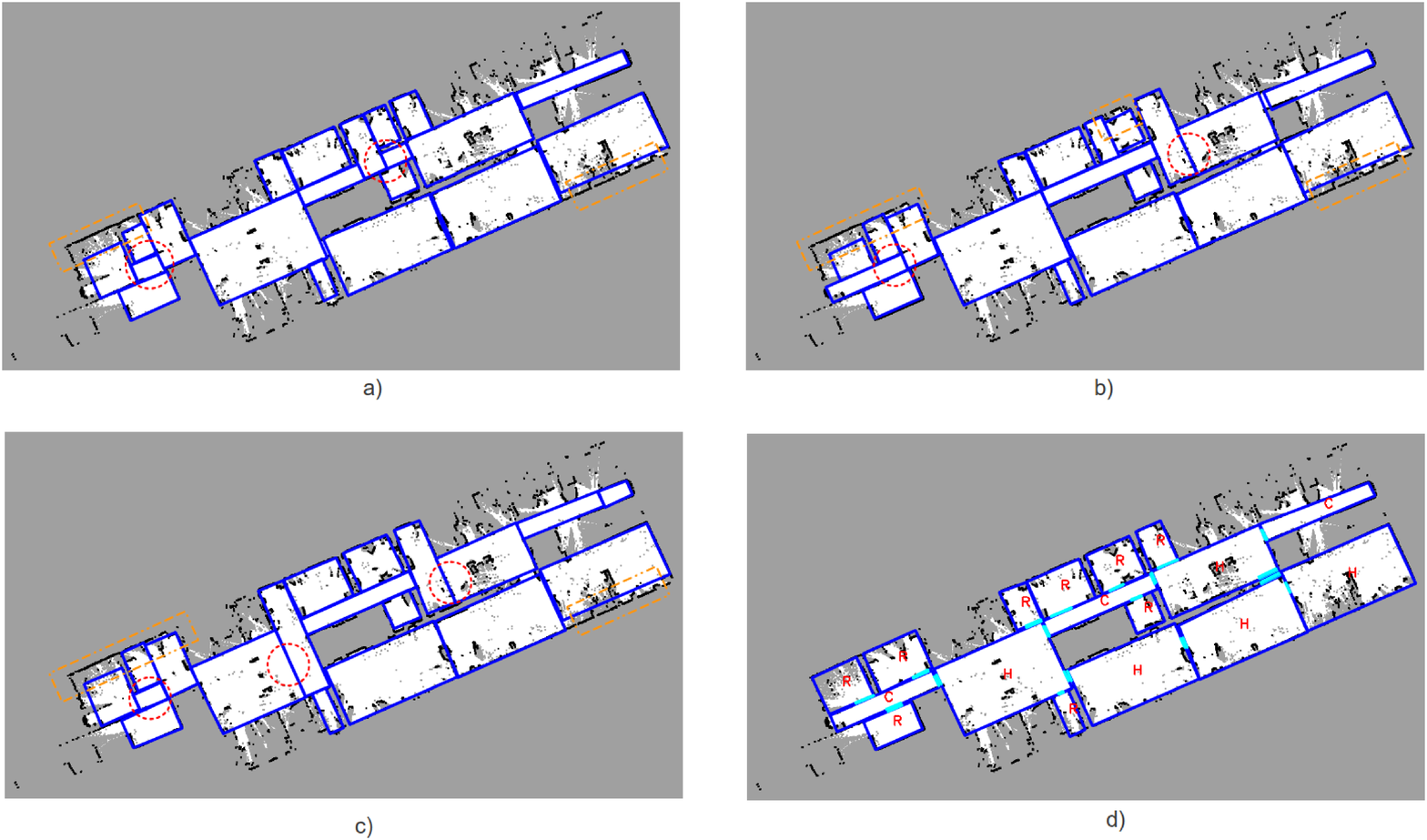}
    \caption{Comparison of overall performance between our previous work \cite{liu2013extracting} (figure a, b and c) and current system (figure d). Structural drawbacks of the previous results are highlighted by red dashed circles, and bad local matches that are improved by better bottom-up feature detectors are highlighted by orange dashed rectangles.}
    \label{figure:comparison1}
\end{figure*}

\begin{figure}[!htb]
	\centering
	\includegraphics[width=.9\columnwidth]{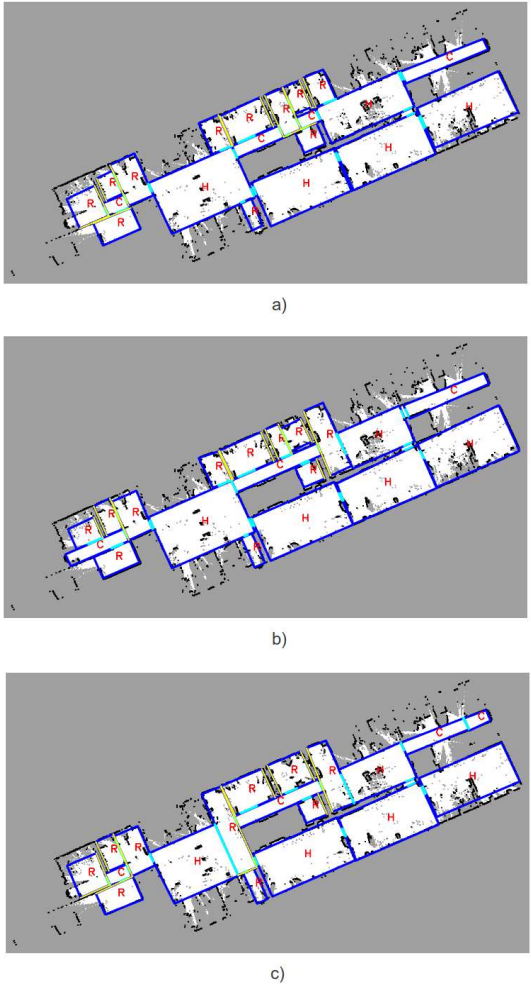}
    \caption{Applying our current system to the results of our previous work. The neighbour walls of adjacent rooms that should have same length are drawn as yellow lines.}
    \label{figure:neighborcheck}
\end{figure}

\indent Compared with our previous work \cite{liu2013extracting}, our current system incorporates task-specific context knowledge in a systematic way, so that the input map can be explained according to the preferred model structure (see Fig. \ref{figure:posterior concept}). A comparison on the overall performance is depicted in Fig. \ref{figure:comparison1}. Since MCMC sampling is a stochastic process, without a corresponding model structure enforced by knowledge processing, the final result can be different for different runs. Three results obtained from our previous work are shown in Fig. \ref{figure:comparison1}-a,b,c. Although all these three results provide good match to the input map (high data likelihood), they have structural drawbacks (low prior) which do not comply with our knowledge, i.e. a human being will not interpret the input map in such ways. By applying our current system to these results, these structural drawbacks can be eliminated so as to generate a semantic world that results in high data likelihood and high prior, i.e. high posterior (worlds shown in Fig. \ref{figure:distribution1}-b). In addition, various bottom-up feature detectors are improved in our current system so that bad local matches (highlighted by orange rectangles in Fig. \ref{figure:comparison1}-a,b,c) can also be corrected. Fig. \ref{figure:neighborcheck} shows the application of our current system to these three results: the units are classified accordingly, and neighbour walls of adjacent rooms are checked for ``same length".

\begin{figure*}[!htb]
	\centering
	\includegraphics[width=1.8\columnwidth]{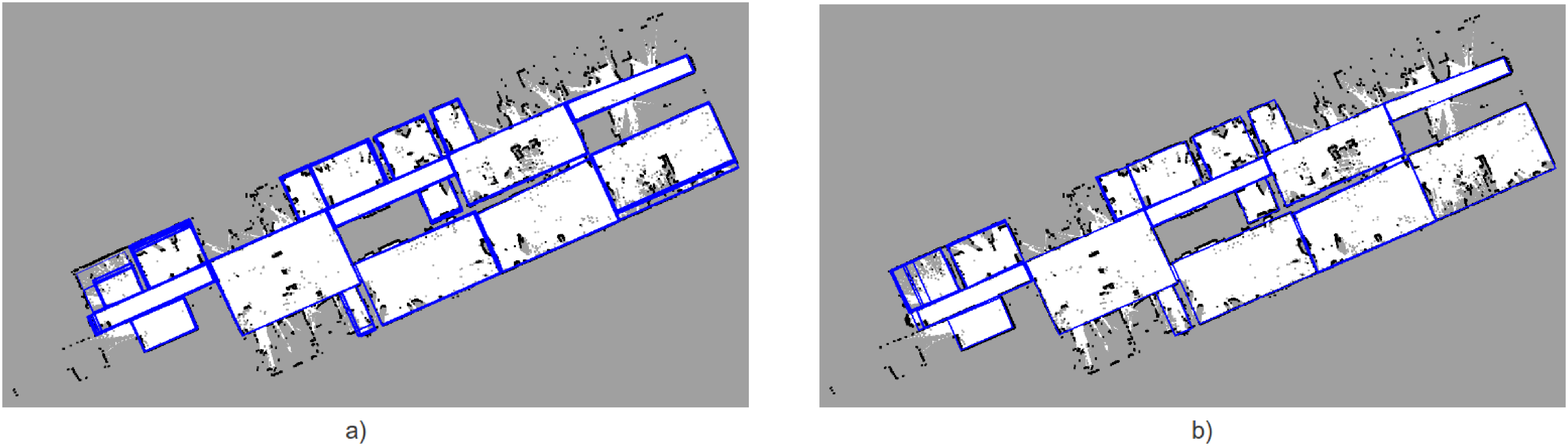}
    \caption{Starting from the world state shown in figure \ref{figure:distribution1}-a, we plot 1000 accepted samples obtained from our previous work (figure a) and from the current system (figure b) onto the classified map . It is obvious that the underlying Markov chain converges better using the current system.}
    \label{figure:distribution}
\end{figure*}

\indent Even using the same semantic world (Fig. \ref{figure:distribution1}-a) as the start state of the Markov chain, the posterior distribution obtained from our previous work and our current system is different. This effect is illustrated in Fig. \ref{figure:distribution} by plotting 1000 accepted worlds together. Here we can see that the semantic worlds obtained from our current system show much smaller variations than the ones obtained from our previous work, indicating that our current system constructs a Markov chain with better stability and better convergence.

\begin{figure*}[!htb]
	\centering
	\includegraphics[width=2\columnwidth, height=.9\columnwidth]{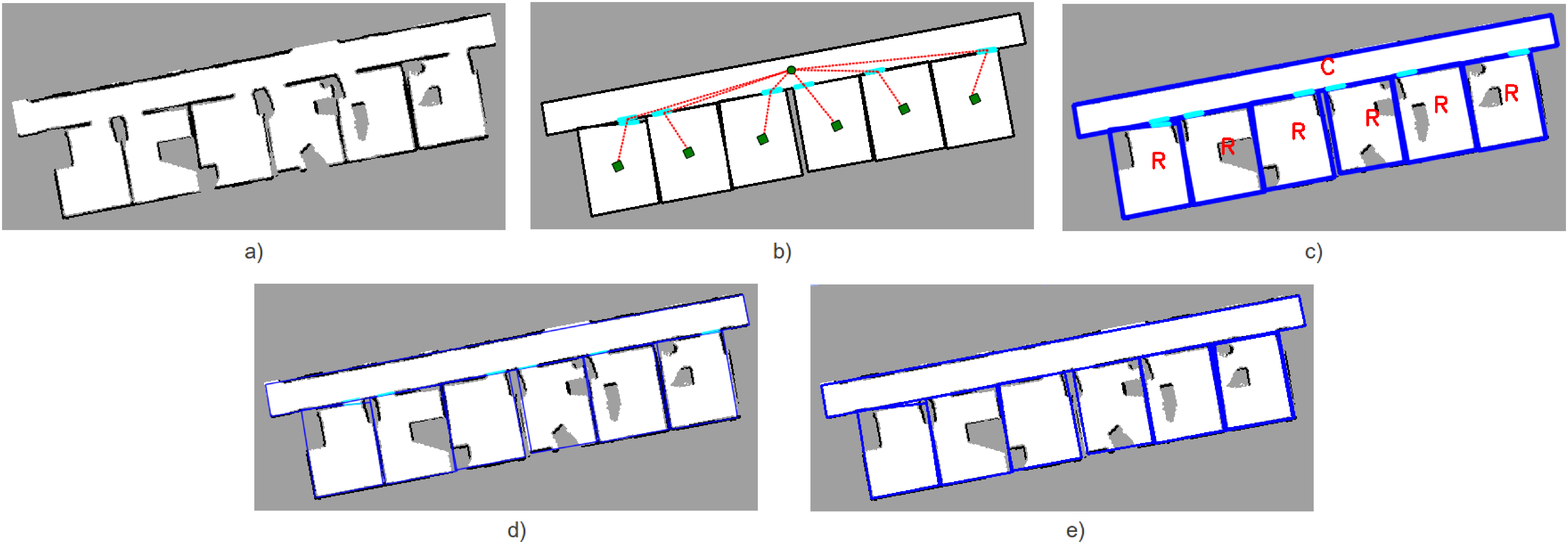}
    \caption{The overall performance of our current system for another data set from \cite{Radish}. Colour code is the same as in Fig. \ref{figure:overall performance}. a) Classified map. b) The corresponding semantic world. c) A direct comparison between the map and the resulting semantic world. d) Drawing the semantic world shown in c) using very thin lines. This figure serves as a comparison with the resulting posterior distribution (built by plotting 1000 accepted samples together) which is depicted in e). }
    \label{figure:intel-overall}
\end{figure*}

\indent Fig. \ref{figure:intel-overall} shows the performance of our current system on another data set obtained from \cite{Radish}. The corresponding posterior distribution after the convergence is depicted in Fig. \ref{figure:intel-overall}-e. Again, we can obviously tell that our system constructs a stable Markov chain that produces a fine semantic world (abstract model) for the input map (data).

\subsection{Evaluation using data acquired by our own robot}
\begin{figure}[H]
	\centering
	\includegraphics[width=.5\columnwidth]{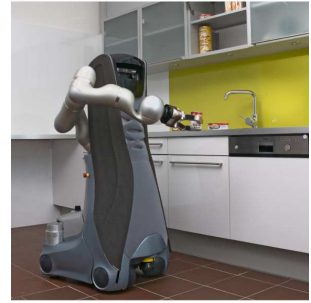}
    \caption{Robots used in real world experiments and simulation. Our mobile robot equipped with three laser scanners, a Kinect camera and a stereo camera system.}
    \label{figure:cob}
\end{figure}

In addition to publicly available benchmark data, we test our system on our own mobile robot (see Fig. \ref{figure:cob}) as well, which is equipped with three laser scanners, a Kinect camera and a stereo camera system. In our experiments, we mainly used the two laser scanners that are situated at the front and the back side of our robot to sense the robot's operating environments. While our robot travels in the environment, the obtained laser scans are fed into the Gmapping algorithm \cite{grisetti2007improved} to generate an occupancy grid map of the perceived environment. Subsequently, the resulting grid map is used as input in our system to produce the corresponding semantic world.

\begin{figure*}[!htb]
	\centering
	\includegraphics[width=1.8\columnwidth]{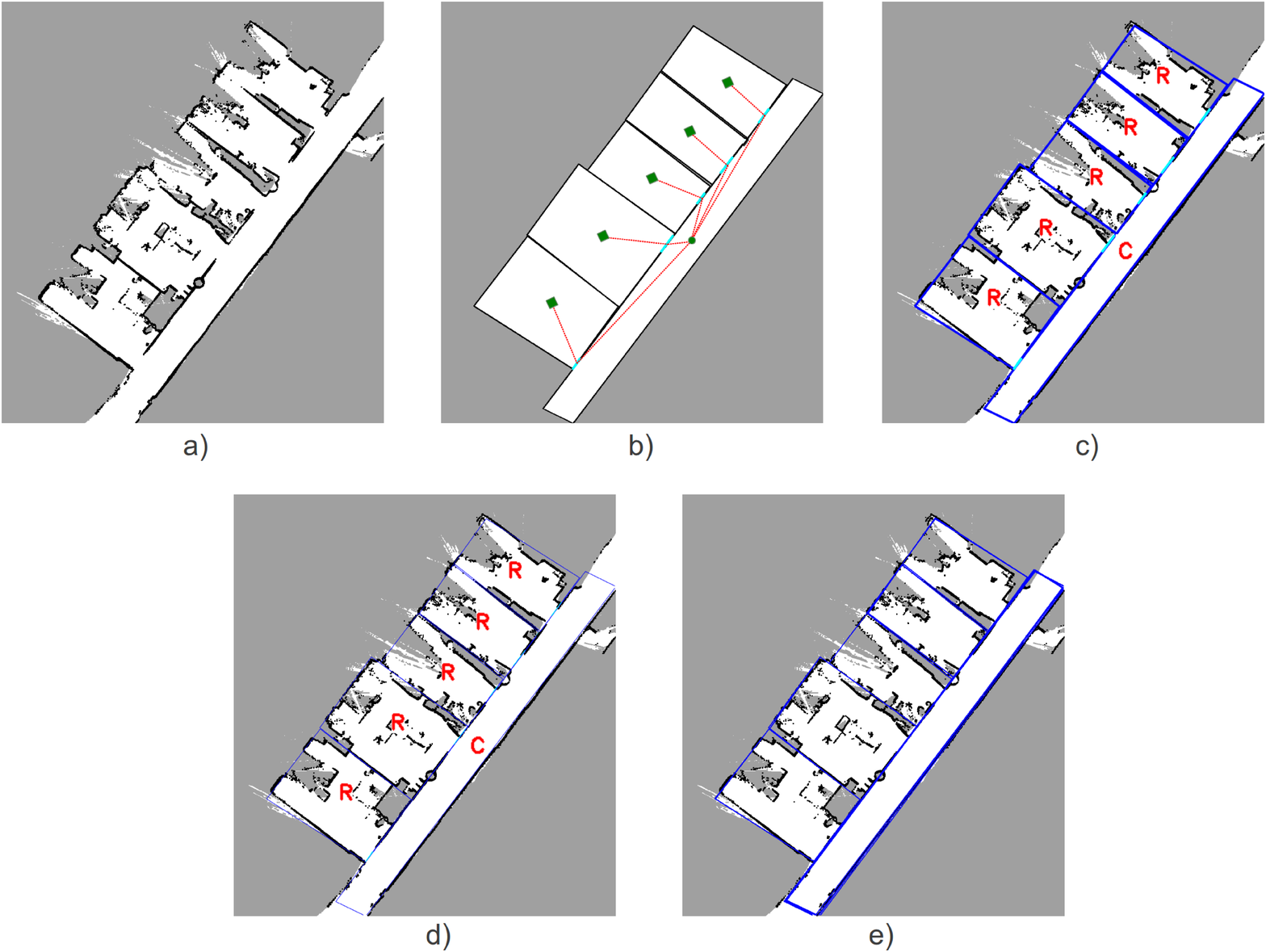}
    \caption{The result of our semantic mapping system using data obtained by our own mobile robot. a) The resulting grid map of a cluttered office environment. b) The semantic world produced by our system. c) A direct comparison between the grid map and the semantic world. d) The same semantic world as shown in c) plotted in very thin line. e) Posterior distribution built by 1000 samples after the underlying Markov chain has reached the state in d).}
    \label{figure:ct535}
\end{figure*}

Fig. \ref{figure:ct535} shows the result of our system for an indoor office environment, which contains five furnished office rooms and a big corridor. In the grid map of this environment (Fig. \ref{figure:ct535}-a), we can see that the five office rooms are quite cluttered (because of the existence of furniture and things). In spite of the clutter, our system still provides a fine semantic world that correctly explains the environment with six space units (five rooms and one corridor) and the correct topology, as shown in Fig. \ref{figure:ct535}-b. In the direct comparison between the grid map and the resulting semantic world, as shown in Fig. \ref{figure:ct535}-c, it is obvious that the resulting semantic world accurately approximates the geometry of the grid map which essentially captures the environment geometry. By plotting 1000 samples together we show the resulting posterior distribution in Fig. \ref{figure:ct535}-e. Here each sample is drawn in very thin line as depicted by Fig. \ref{figure:ct535}-d. Again, we can obviously see that our system constructs a stable Markov chain that well converges to the goal distribution.

\subsection{Evaluation using simulated data obtained from open source simulators}
\begin{figure}[!htb]
	\centering
	\includegraphics[width=1.0\columnwidth]{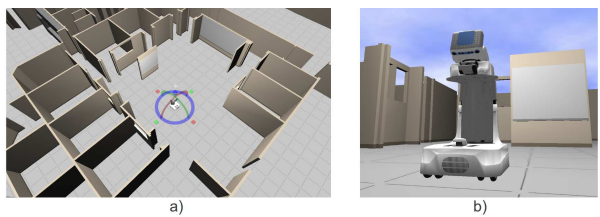}
    \caption{An example of the simulated PR2 robot and its operating environment. a) A simulated 3D environment. b) A simulated PR2 robot.}
    \label{figure:pr2-simulation}
\end{figure}

In addition to the above experiments with real world data, we evaluate our system in simulation as well. In the real world, we do not encounter so many indoor environments of different structures, in which we can test our system. Thus it is quite helpful to evaluate our system in simulation where a big number of different environments can be manually created. Without loss of generality, we use open source simulators and 3D environment models for this purpose, which are publicly available in the internet. Here we have used the ROS \cite{ros} integration of the Gazebo simulator \cite{gazebo} to simulate a PR2 robot \cite{pr2} and its operating environments. An example of this robot and a simulated 3D environment is depicted by Fig. \ref{figure:pr2-simulation}.

\begin{figure*}[!htb]
	\centering
	\includegraphics[width=2\columnwidth]{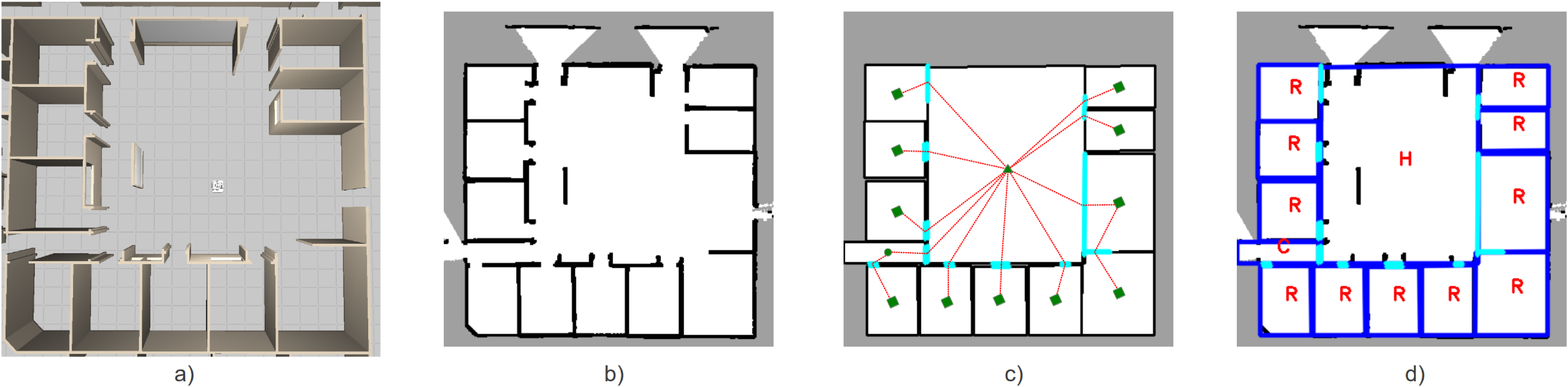}
    \caption{Simulation result no. 1. a) A snapshot of the simulated environment in the Gazebo simulator. b) The corresponding grid map generated by the Gmapping algorithm. c) The resulting semantic world with its topology obtained by our system. Small triangles, circles and rectangles show the geometric center of halls, corridors and rooms. d) A direct comparison between the grid map and the semantic world.}
    \label{figure:wg1}
\end{figure*}

\begin{figure*}[!htb]
	\centering
	\includegraphics[width=2\columnwidth]{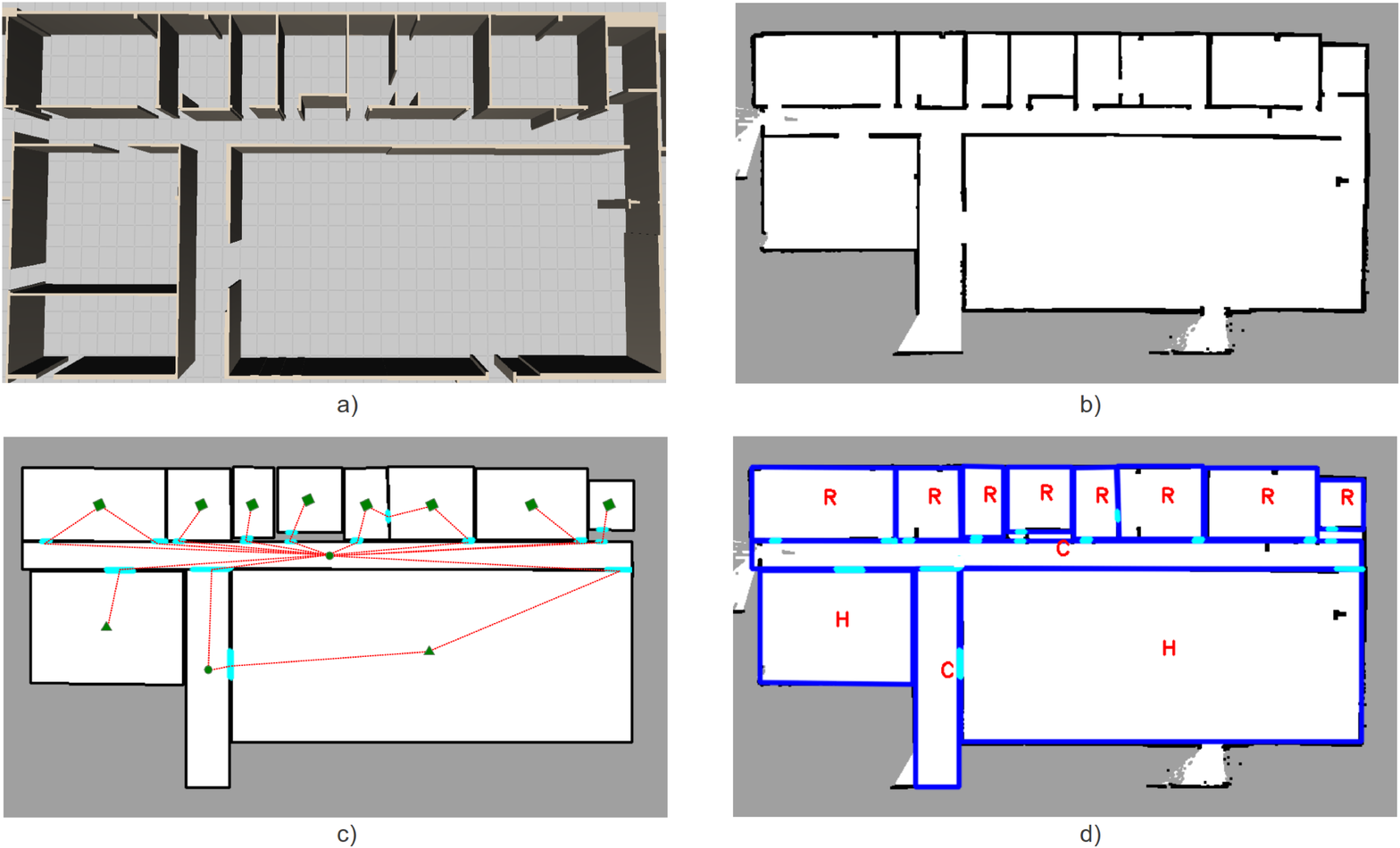}
    \caption{Simulation result no. 2. a) A snapshot of the simulated environment in the Gazebo simulator. b) The corresponding grid map generated by the Gmapping algorithm. c) The resulting semantic world with its topology obtained by our system. Small triangles, circles and rectangles show the geometric center of halls, corridors and rooms. d) A direct comparison between the grid map and the semantic world.}
    \label{figure:wg2}
\end{figure*}

\begin{figure*}[!htb]
	\centering
	\includegraphics[width=2\columnwidth]{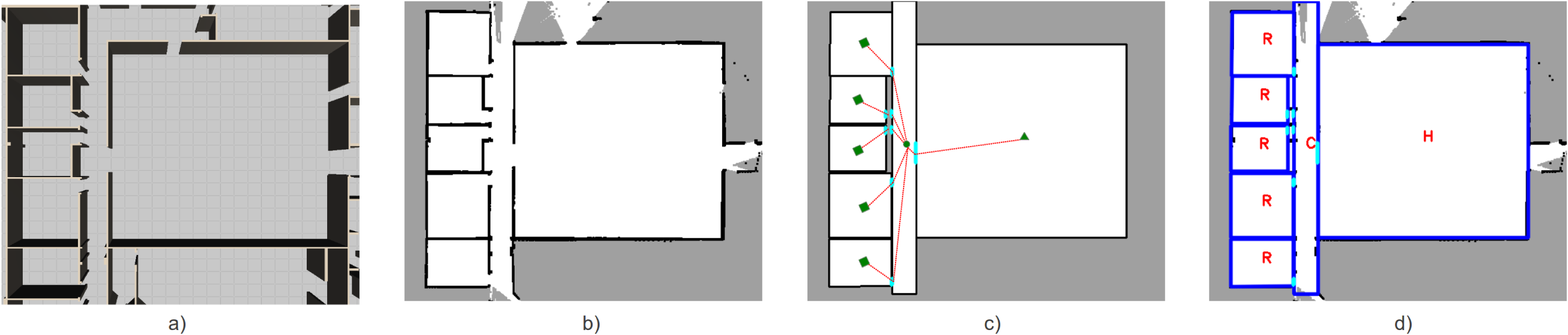}
    \caption{Simulation result no. 3. a) A snapshot of the simulated environment in the Gazebo simulator. b) The corresponding grid map generated by the Gmapping algorithm. c) The resulting semantic world with its topology obtained by our system. Small triangles, circles and rectangles show the geometric center of halls, corridors and rooms. d) A direct comparison between the grid map and the semantic world.}
    \label{figure:wg3}
\end{figure*}

\begin{figure*}[!htb]
	\centering
	\includegraphics[width=2\columnwidth]{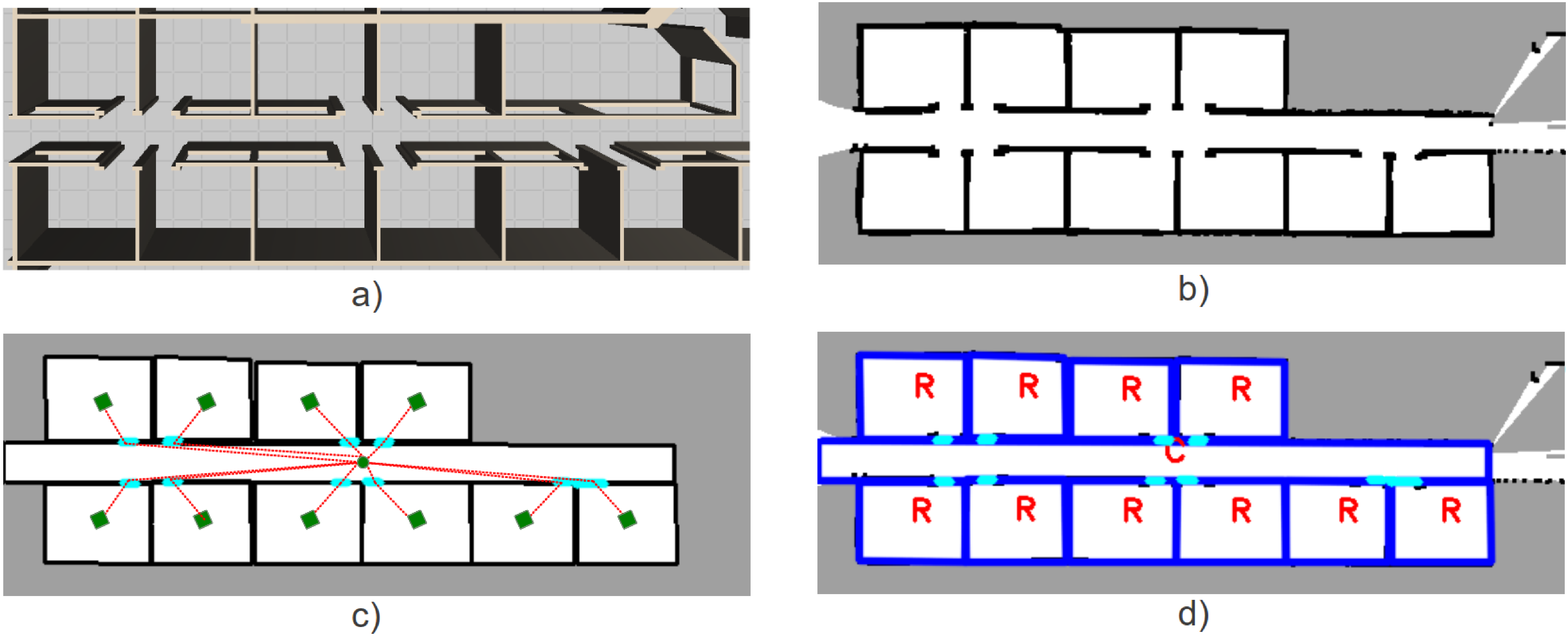}
    \caption{Simulation result no. 4. a) A snapshot of the simulated environment in the Gazebo simulator. b) The corresponding grid map generated by the Gmapping algorithm. c) The resulting semantic world with its topology obtained by our system. Small triangles, circles and rectangles show the geometric center of halls, corridors and rooms. d) A direct comparison between the grid map and the semantic world.}
    \label{figure:wg4}
\end{figure*}

\begin{figure*}[!htb]
	\centering
	\includegraphics[width=2\columnwidth]{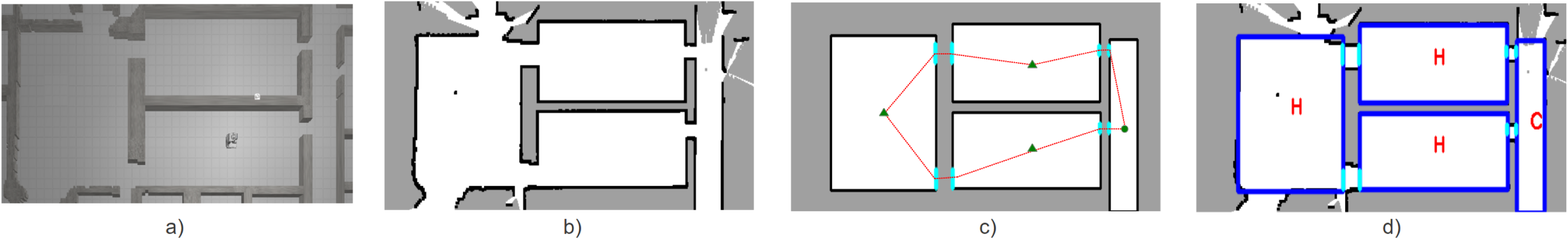}
    \caption{Simulation result no. 5. a) A snapshot of the simulated environment in the Gazebo simulator. b) The corresponding grid map generated by the Gmapping algorithm. c) The resulting semantic world with its topology obtained by our system. Small triangles, circles and rectangles show the geometric center of halls, corridors and rooms. d) A direct comparison between the grid map and the semantic world.}
    \label{figure:wg5}
\end{figure*}

Fig. \ref{figure:wg1} to Fig. \ref{figure:wg5} show five simulation results. In these figures, sub-figures a) show snapshots of the 3D environments simulated by the Gazebo simulator. Sub-figures b) depict the corresponding grid maps generated by the Gmapping algorithm, after the simulated robot has perceived the environments. Sub-figures c) illustrate the resulting semantic worlds with their topology, where the geometric centers of halls, corridors and rooms are shown by small triangles, circles and rectangles respectively. Finally, a direct comparison is shown in sub-figures d) by plotting the semantic worlds onto the corresponding grid maps. 

We purposefully chose these five environments to test our system, because they represent several common environment types which are often found in the reality. The environment shown in Fig. \ref{figure:wg1} represents the type, in which a big hall is surrounded by a lot of satellite rooms. Fig. \ref{figure:wg2} depicts a complex indoor environment consisting of many space units. In this environment, rooms are located in a row and are connected by corridors, which separate halls from rooms. Another environment of this kind is illustrated in Fig. \ref{figure:wg3}. Fig. \ref{figure:wg4} shows a classical office environment, where ten rooms are situated in two rows and connected by a long corridor. Another environment comprised of three halls and a corridor, which is like an exhibition centre, is depicted in Fig. \ref{figure:wg5}. As we can see, our system performs very well in all the five environments, i.e. the resulting semantic world well explains the corresponding environment with a correct number of space units and an appropriate topology. Moreover, as the quantitative evaluation in the following sub-section will show, the resulting semantic worlds accurately represent the geometry of the perceived environments as well.

\subsection{Quantitative evaluation}
\indent In order to quantitatively evaluate our approach, we compute $K(W,M)$, the \emph{cell prediction rate} capturing the predictive power of the semantic world model $W$ with respect to an input map $M$:
\begin{equation}
K(W,M)=\frac{\sum\limits_{c(x,y)\in M }l(c(x,y))}{t_M} \nonumber,
\end{equation}
with
\begin{equation}
l(c(x,y))=\left\{\begin{array}{lcc}
1,~~ C_M(x,y)=C_W(x,y),\\
0,~~ \textrm{otherwise},
\end{array}
\right.
\end{equation}
where $t_M$ is the number of all grid cells in the map $M$. $c(x,y)$ indicates one grid cell located at the position $(x,y)$. $C_M(x,y)$ and $C_W(x,y)$ are previously defined in equation (\ref{equ:classify}) and (\ref{equ:classify2}). $K(W,M)$ of the maps shown in this paper is given in Table \ref{TAB:result}. In this table, we can see that the $K(W,M)$ for the three real world data sets (Fig. \ref{figure:overall performance}-d, Fig. \ref{figure:intel-overall}-c and Fig. \ref{figure:ct535}-c) is above 90\%. The mismatch is mainly due to the clutter caused by furniture and things. For the five simulation data sets (Fig. \ref{figure:wg1}-d, Fig. \ref{figure:wg2}-d, Fig. \ref{figure:wg3}-d, Fig. \ref{figure:wg4}-d and Fig. \ref{figure:wg5}-d), the $K(W,M)$ is above 94\%, where the mismatch lies mainly in some not-fully-explored areas. Such areas are evidence for partially explored space units in corresponding environments but are too small to be recognized. To sum up, our system accurately represents the geometry of the perceived environments in all experiments.

\begin{table*}[htb]
	\centering
	\caption{Cell prediction rate $K(W,M)$.}
	\begin{tabular}{|c||*{5}{c|c|c|c|c|c|c|c|}}\hline
	 	&\textbf{Fig.}~\ref{figure:overall performance}-d	&\textbf{Fig.}~\ref{figure:intel-overall}-c&\textbf{Fig.}~\ref{figure:ct535}-c&\textbf{Fig.}~\ref{figure:wg1}-d&\textbf{Fig.}~\ref{figure:wg2}-d&\textbf{Fig.}~\ref{figure:wg3}-d&\textbf{Fig.}~\ref{figure:wg4}-d&\textbf{Fig.}~\ref{figure:wg5}-d\\\hline
	Percentage	&93.6\%&91.0\%&90.1\%	&95.4\%&96.0\%&95.2\%&96.3\%&94.4\%\\\hline
	\end{tabular}
	\label{TAB:result}
\end{table*}


\section{Summary and Outlook}
\label{sum}
\indent In this paper, we propose a generalizable knowledge framework for data abstraction, i.e. finding compact abstract model for input data using predefined abstract terms. Based on these abstract terms, intelligent autonomous systems, such as a robot, should be able to make inference according to specific knowledge base, so that they can better handle the complexity and uncertainty of the real world. We propose to realize this framework by combining Markov logic networks (MLNs) and data driven MCMC sampling, because the former are a powerful tool for modelling uncertain knowledge and the latter provides an efficient way to draw samples from unknown complex distributions. Furthermore, we show in detail how to adapt this framework to a certain task, in particular, semantic robot mapping. Based on MLNs, we formulate task-specific context knowledge as descriptive soft rules which increase the overall abstraction performance. Experiments using real world data and simulated data show promising results and thus confirm the usefulness of our framework.

\indent At the current stage, we focus on extracting semantic model from 2D map data, an extension to 3D scenarios is planed. In addition, we plan to improve the performance of our system by applying a probabilistic classifier (currently deterministic) and a more advanced knowledge base. A second line of research will concentrate on applying our framework to other domains.

\section*{Acknowledgements}

This work is accomplished with the support of the Technische Universit\"at M\"unchen - Institute for Advanced Study, funded by the German Excellence Initiative.

The input maps shown in Fig.~\ref{figure:overall performance} and Fig.~\ref{figure:intel-overall} were obtained from the Robotics Data Set Repository (Radish)~\cite{Radish}. Thanks go to Cyrill Stachniss and Andrew Howard for providing these datasets.

We thank Michael Fiegert for his valuable discussions and suggestions.




\bibliographystyle{elsarticle-num}
\bibliography{overall-bibliography}




%
%
%

\break

\vspace{1cm}
\begin{wrapfigure}{l}{1.5cm}
\includegraphics[width=2cm]{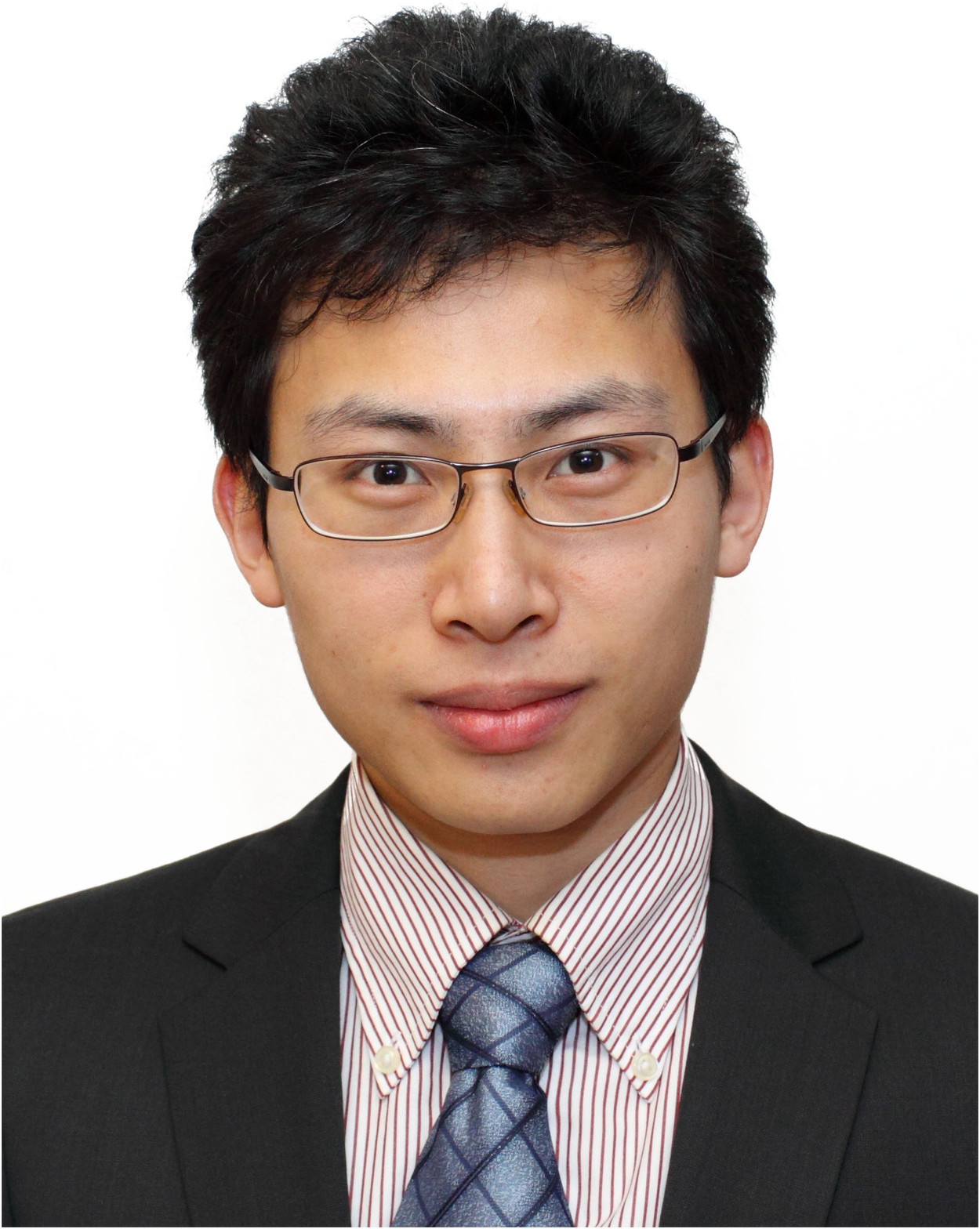}
\end{wrapfigure}
\small \noindent\textbf{Ziyuan Liu} received his B.E. degree in Mechatronics from the TongJi University, Shanghai, China, in 2008. He received his M.S. degree in 2010 from the Institute of Automatic Control Engineering at Technische Universit\"at M\"unchen, Munich, Germany, where he is a Ph.D. candidate currently. His research interests are semantic perception and sampling based inference methods.\\

\begin{wrapfigure}{l}{1.5cm}
\includegraphics[width=2cm]{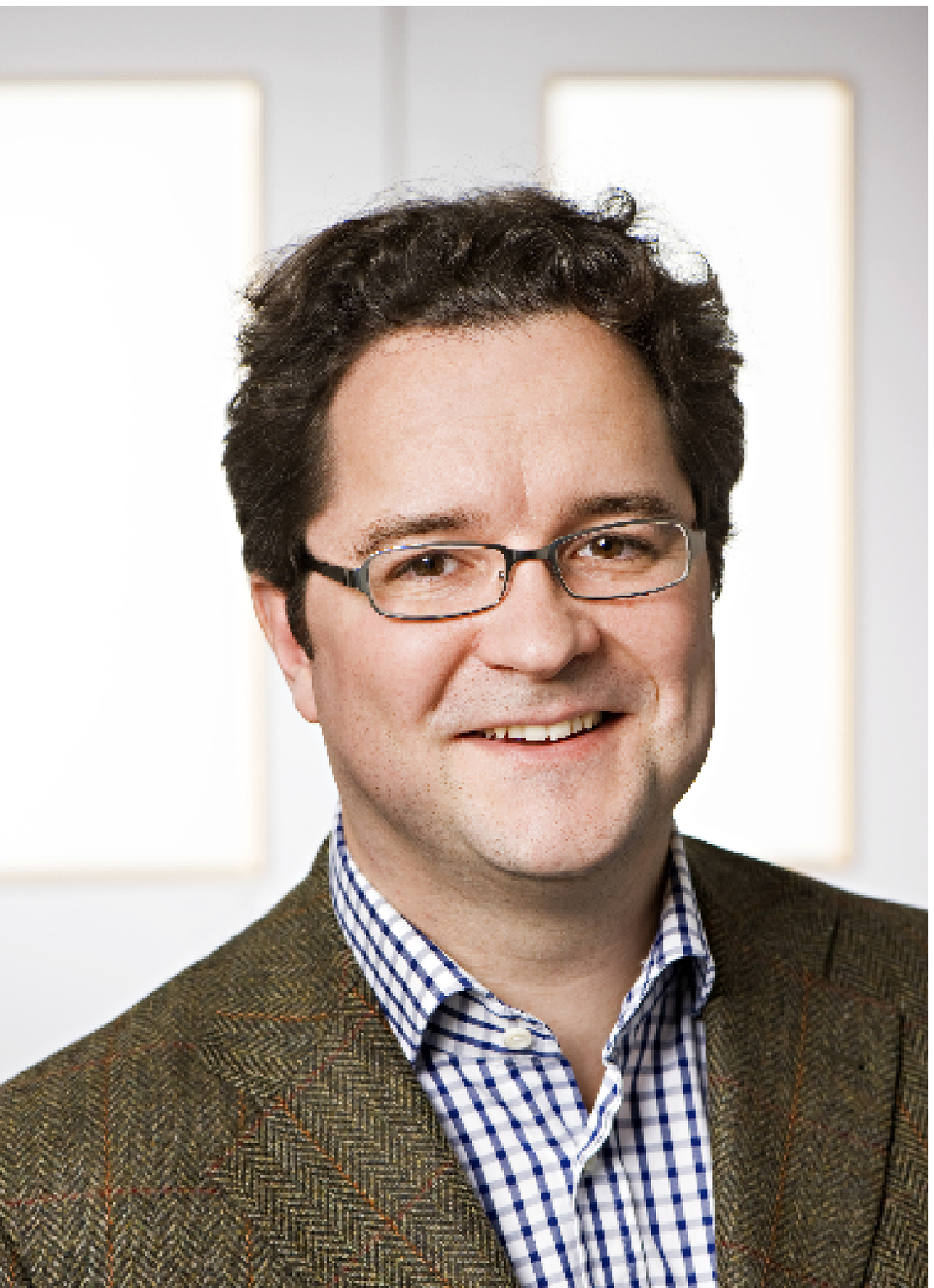}
\end{wrapfigure}

\small \noindent\textbf{Georg von Wichert} received his Diploma (MSc) in Electrical and Control Engineering from Darmstadt University of Technology in 1992. From 1992 to 1998 he was a research and teaching assistant at the Institute of Control Engineering at Darmstadt University of Technology. In Darmstadt he also received the Ph.D. degree in Electrical Engineering in 1998. Since 1998 he is with Siemens Corporate Technology. At the same time he is a fellow of the Institute for Advanced Study at Technische Universit\"at M\"unchen.
\end{document}